\documentclass{article}

\usepackage{graphicx}
\usepackage{subfigure}

\newcommand{\comment}[1]{}
\usepackage{amsmath}
\usepackage{amsthm}
\usepackage{bm}
\usepackage{comment}
\usepackage{xcolor}
\usepackage{soul}

\usepackage{algorithm}
\usepackage{algorithmic}

\newtheorem{theorem}{Theorem}[section]
\newtheorem{lemma}{Lemma}[section]
\newtheorem{corollary}{Corollary}[section]

\newtheorem{remark}{Remark}[section]

\usepackage{xspace}
\usepackage[inline]{enumitem}
\newenvironment{denseitemize}{
\begin{itemize}[topsep=2pt, partopsep=0pt, leftmargin=1.5em]
  \setlength{\itemsep}{2pt}
  \setlength{\parskip}{0pt}
  \setlength{\parsep}{0pt}
}{\end{itemize}}

\usepackage[utf8]{inputenc} 
\usepackage[T1]{fontenc}    
\usepackage{hyperref}       
\usepackage{url}            
\usepackage{booktabs}       
\usepackage{amsfonts}       
\usepackage{nicefrac}       
\usepackage{microtype}      
\usepackage{xcolor}         
\def\name{Fed-ensemble\xspace}
\title{\name : Improving Generalization through Model Ensembling in Federated Learning}

\author{%
  Naichen Shi \thanks{IOE, University of Michigan},
  Fan Lai \thanks{CSE, Univeristy of Michigan},
  Raed Al Kontar \thanks{IOE, University of Michigan},
  Mosharaf Chowdhury \thanks{CSE, University of Michigan}

}

\begin{document}

\maketitle

\begin{abstract}
  In this paper we propose \name: a simple approach that brings model ensembling to federated learning (FL). Instead of aggregating local models to update a single global model, \name uses random permutations to update a group of $K$ models and then obtains predictions through model averaging. Fed-ensemble can be readily utilized within established FL methods and does not impose a computational overhead as it only requires one of the $K$ models to be sent to a client in each communication round. Theoretically, we show that predictions on new data from all $K$ models belong to the same predictive posterior distribution under a neural tangent kernel regime. This result in turn sheds light on the generalization advantages of model averaging. We also illustrate that Fed-ensemble has an elegant Bayesian interpretation. Empirical results show that our model has superior performance over several FL algorithms, on a wide range of data sets, and excels in heterogeneous settings often encountered in FL applications.
\end{abstract}
\section{Introduction}
The rapid increase in computational power on edge devices has set forth federated learning (FL) as an elegant alternative to traditional cloud/data center based analytics. FL brings training to the edge, where devices collaboratively extract knowledge and learn complex models (most often deep learning models) with the orchestration of a central server while keeping their personal data stored locally. This paradigm shifts not only reduces privacy concerns but also sets forth many intrinsic advantages including cost efficiency, diversity, and reduced communication, amongst many others \cite{yang2019federated,kairouz2019advances}.

The earliest and perhaps most popular FL algorithm is FederatedAveraging (fedavg) \cite{fedavg}. In fedavg, the central server broadcasts a global model (set of weights) to selected edge devices, these devices run updates based on their local data, and the server then takes a weighted average of the resulting local models to update the global model. This process iterates over hundreds of training rounds  to maximize the performance for all devices.  Fedavg has seen prevailing empirical successes in many real-world applications \cite{servicefl,hard2018federated}. The caveat, however, is that aggregating local models is prone to overfitting and suffers from high variance in learning and prediction when local datasets are heterogeneous be it in size or distribution or when clients have limited bandwidth, memory, or unreliable connection that effects their participation in the training process  \cite{nishio2019client, moreno2012unifying}. Indeed, in the past few years, multiple papers have shown the high variance in the performance of FL algorithms and their vulnerability to overfitting,  specifically in the presence of data heterogeneity or unreliable devices \cite{jiang2019improving, wang2019federated, smith2017federated, nishio2019client, moreno2012unifying}. Since then some notable algorithms have attempted to improve the generalization performance of fedavg and tackle the above challenges as discussed in Sec. \ref{sec:literature}. 

In this paper, we adopt the idea of ensemble training to FL and propose \textit{\name} which iteratively updates an ensemble of models to improve the generalization performance of FL methods. We show, both theoretically and empirically, that ensembling is efficient in reducing variance and achieving better generalization performance in FL. Our approach is orthogonal to current efforts in FL aimed at reducing communication cost, heterogeneity or finding better fixed points as such approaches can be directly integrated into our ensembling framework. Specifically, we propose an algorithm to train $K$ different models. Predictions are then obtained by ensembling all trained modes, i.e. model averaging. Our contributions are summarized below:
\begin{itemize}
\itemsep0em 
\item \textbf{Model:} 
We propose an ensemble treatment to FL which updates $K$ models over local datasets. Predictions are obtained by model averaging. Our approach does not impose additional burden on clients as only one model is assigned to a client at each communication round. We then show that \textit{\name} has an elegant Bayesian interpretation. 

\item \textbf{Convergence and Generalization:} We motivate the generalization advantages of ensembling under a bias-variance decomposition. Using neural tangent kernels (NTK), we then show that predictions at new data points from all $K$ models converge to samples from the same limiting Gaussian process in sufficiently overparameterized regimes. This result in turn highlights the improved generalization and reduced variance obtained from averaging predictions as all $K$ models converge to samples from that same limiting posterior. To the best of our knowledge, this is the first theoretical proof for the convergence of general multilayer neural network in FL in the kernel regime, as well as the first justification for using model ensembling in FL. Our proof also offers standalone value as it extends NTK results to FL settings.

\item \textbf{Numerical Findings:} 
We demonstrate the superior performance of \textit{\name} over several FL techniques on a wide range of datasets, including realistic FL datasets in FL benchmarks~\cite{fedscale}. Our results highlight the capability of \textit{\name} to excel in heterogeneous data settings. 
\end{itemize}

\section{Related Work} 
\label{sec:literature}


\paragraph{Single-mode FL} 
Many approaches have been proposed to tackle the aformentioned FL challenges. Below we list a few, yet this is by no means an extensive list. \textit{Fedavg} \cite{fedavg} allows inexact local updating to balance communication vs. computation on resource-limited edge devices, while reporting decent performance on mitigating performance bias on skewed client data distributions. 
\textit{Fedprox} \cite{fedprox} attempts to solve the heterogeneity challenge by adding a proxy term to control the shift of model weights between the global and local client model. This proxy term can be  viewed as a normal prior distribution on model weights. 
There are several influential works aiming at expediting convergence, like \textit{FedAcc} \cite{fedacc}, \textit{FedAdam} and \textit{FedYogi} \cite{fedadam}, reducing communication cost, like \textit{DIANA} \cite{diana}, \textit{DORE} \cite{dore}, or find better fixed points, like \textit{FedSplit} \cite{fedsplit}, \textit{FedPD} \cite{fedpd}, \textit{FedDyn} \cite{feddyn}. \emph{As aformentioned, these efforts are complementary to our work, and they can be integrated into our ensembling framework}. 

\textbf{Ensemble of deep neural nets} Recently ensembling methods in conventional, non-federated, deep learning has seen great success. Amongst them, \cite{fastensembling} analyzes the loss surface of deep neural networks and uses cyclic learning rate to learn multiple models and form ensembles. \cite{deepensemble} visually demonstrates the diversity of randomly initialized neural nets and empirically shows the stability of ensembled solutions.  Also, \cite{multiswag} connects ensembling to Bayesian deep neural networks and highlights the benefits of ensembling. 

\textbf{Bayesian methods in FL and beyond}
There are also some recent attempts to exploit Bayesian philosophies in FL. Very recently, \textit{Fedbe} was proposed \cite{fedbe} to couple Bayesian model averaging with knowledge distillation on the server. Fedbe formulates the Gaussian or Dirichlet distribution of local models and then uses knowledge distillation on the server to update the global model. This procedure however requires additional datasets on the server and a significant computational overhead, thus being demanding in FL. Besides that, Bayesian non-parametrics have been investigated for advanced model aggregation through matching and re-permuting neurons using neuron matching algorithms \cite{fedma, bnfed}. Such approaches intend to address the permutation invariance of parameters in neural networks, yet suffer from a large computational burden. We also note that Bayesian methods have been also exploited in meta-learning which can achieve personalized FL. For instance,  \cite{bayesianmaml} proposes a Bayesian version of MAML \cite{finn2017model} using Stein variational gradient descent \cite{svgd}. An amortized version of this  \cite{amortizedmaml} utilizes variational inference (VI) across tasks to learn a prior distribution over neural network weights.

\section{\name}
\label{sec:algorithm}


\subsection{Parameter updates through Random Permutation}

Let $N$ denote the number of clients where the local dataset of client $i$ is given as $\bm{D}_i=\{x_{ij},y_{ij}\}_{j\in [1,2,.., n_i]}$ and $n_i$ is the number of observations for client $i$. Also, let $\bm{D}$ be the union of all local datasets $\bm{D}=\bigcup_{i=1}^N\bm{D}_i$, and $f_{\bm{w}}(\cdot)$ denote the model to be learned parametrized by weight vector $\bm{w}$.

\paragraph{Design principle}
Our goal is to get multiple models ($f_{\bm{w}_1}, f_{\bm{w}_2}, .., f_{\bm{w}_K})$ engaged in the training process. Specifically, we use $K$ models in the ensemble, where $K$ is a predetermined number. The $K$ models are randomly initialized by standard initialization, e.g. Xavier \cite{xavierinitialization} or He \cite{heinitialization}. In each training round, \textit{\name} assigns one of the $K$ models to the selected clients to train on their local data. The server then aggregates the updated models from the same initialization. All $K$ models eventually learn from the entire dataset, and allow an improvement to the overall inference accuracy by averaging over $K$ predictions produced by each model. Hereon, we use mode or weight to refer to ${\bm{w}_k}$ and model to refer to the corresponding $f_{\bm{w}_k}$.

\paragraph{Objective of ensemble training}
Since we aim to learn $K$ modes, the objective of FL training can be simply defined as:
\begin{equation}
\label{eqn::objective}
 \min_{\mathcal{W}}\sum_{k=1}^K\sum_{i=1}^Np_i\ell_i(\bm{w}_k)
\end{equation}
where $\mathcal{W}=\{\bm{w}_1, \bm{w}_2, ..., \bm{w}_K\}$, $\ell_i$ is the local empirical loss on client $i$,  
$\ell_i\left(\bm{w}_k\right)=\\ \frac{1}{n_i}\sum_{(x_{ij},y_{ij})\in\bm{D}_i}L(f_{\bm{w}_k}(x_{ij}),y_{ij})$, $p_i$ is a weighting factor for each client and $L$ is a loss function such as cross entropy. 

\begin{algorithm}[h]
\centering
   \caption{\name: \name using random permutations}
   \label{alg::bfl}
\begin{algorithmic}[1]
   \STATE {\bfseries Input:} Client datasets $\{\bm{D}_i\}_{i=1}^N$, $T$, $K$, $M$,  initialization for $\{\bm{w}_k\}_{k=1}^K$ 
   
   \STATE Randomly divide $N$ clients into $Q$ strata  $\{S_q\}_{q=1}^Q$ 
   
   \FOR{$t$ = 1, 2, $\cdots T$}
   
   \STATE Index Permutation:  $\bm{P}_{q, \cdot}$ = \textit{shuffle\_list$[1,2,\cdots K]$} 
   
   \FOR {$r$ = 1, 2, $\cdots K$}
   \FOR {$q$ = 1, 2, $\cdots Q$}
   
   \STATE Randomly select clients $C_q$ from $S_q$ 
   
   \STATE Server broadcasts mode $\bm{w}_{\mathbf{P}_{q,r}}$ to clients $C_q$ 
   
   \FOR{each client $i$ in $C_q$} 
   
   \STATE $\bm{w}^{(i)}=local\_training [\bm{w}=\bm{w}_{\mathbf{P}_{q,r}}, \bm{D}_i]$
   
   \STATE Client $i$ sends $\bm{w}_i$ to server
   \ENDFOR
   \ENDFOR
   \STATE \textit{server\_update} [$\bm{w}^{(1)}, .., \bm{w}^{(M)}$]. 
   \ENDFOR
   
   \ENDFOR
\end{algorithmic}
\end{algorithm}

\paragraph{Model training with \name}
We now introduce our algorithm \textit{\name} (shown in  Algorithm \ref{alg::bfl}), which is inspired by random permutation block coordinate descent used to optimize \eqref{eqn::objective}. 
We first, randomly divide all clients into $Q$ strata, $\{S_q\}_{q=1}^Q$. Now let $r \in [1, .., K]$ denote an individual communication round where $M$ clients are sampled in each round. Also, let $t\in[1,.., T]$ denote the training age, where each age consists of $K$ communication rounds. Hence the total number of communication rounds for $T$ ages is $R= T \times K$. At the beginning of each training age, every stratum will decide the order of training in this age. To do so, define a permutation matrix $\bm{P}$ of size $Q\times K$ such that at each age $t$ the rows of $\bm{P}$ are given as $\bm{P}_{q, \cdot} = {\textit{shuffle\_list}[1,2,\cdots K]}$. More specifically, in the $r$-th communication rounds in this age, the server samples some clients from each stratum. Clients from stratum $q$ get assigned mode $\bm{w}_{\mathbf{P}_{q,r}}$ as their initialization for $\bm{w}_i$ and then do a training procedure on their local data, denoted as $\bm{w}^{(i)}=local\_training [\bm{w}=\bm{w}_{\mathbf{P}_{q,r}}, \bm{D}_i]$. Note that the use of the random permutation matrix $\bm{P}$, not only ensures that modes are trained on diverse clients but also guarantees that every mode is downloaded and trained on all strata in one age. Upon receiving updated models from all $M$ clients, the server 
activates \textit{server\_update} to calculate the new [$\bm{w}^{(1)}, .., \bm{w}^{(M)}$].

\begin{remark}
The simplest form of the \textit{server\_update} function is to average modes from the same initialization: $
\label{eqn::serverupdatesimple}
\bm{w}_k\leftarrow\frac{\sum_{i=1}^M\delta_{ik}\bm{w}^{(i)}}{\sum_{i=1}^M\delta_{ik}}\ \forall k
$,
where $\delta_{ik}=1$ if client $i$ downloads model $k$ at communication round $r$, and 0 otherwise. $\delta_{ik}$ can also be obtained from $\bm{P}$. This approach is an extension of fedavg. However, one can directly utilize any other scheme in \textit{server\_update} to aggregate modes from the same initialization. Similarly different local training schemes can be used within \textit{local\_training}. 
\end{remark}

\begin{remark} The use of multiple modes does not increase computation or communication overhead on clients compared with single-mode FL algorithms, as for every round, only one mode is sent to and trained on each client. However, after carefully selecting modes, models trained by \name can efficiently learn statistical patterns on all clients as proven in the convergence and generalization results in Sec. \ref{sec:theory}.
\end{remark}

\textbf{Model prediction with \name}: After the training process is done, all $K$ modes are sent to the client, and model prediction at a new input point $x^\star$ is achieved via ensembling. 

\begin{center}
    \noindent\fcolorbox{white}[rgb]{0.95,0.95,0.95}{\begin{minipage}{0.6\columnwidth}
		\begin{center}
$f_\mathcal{W}(x^\star)= \frac{1}{K}\sum_{k=1}^K f_{\bm{w}_k}(x^\star) \notag$
\end{center}
\end{minipage}}
\end{center}


\subsection{Bayesian Interpretation of \name} 
\label{sec::bayesianinterpretation}
Interestingly, \textit{\name} has an elegant Bayesian interpretation as a variational inference (VI) approach to estimate a posterior Gaussian mixture distribution over model weights. To view this, let model parameters $\bm{w}$ admit a posterior distribution $p(\bm{w}|\bm{D})$. Under VI, a variational distribution $q(\bm{w})$ is used to approximate $p(\bm{w}|\bm{D})$ by minimizing the KL-divergence between the two: \begin{equation}
\label{eqn::kldivdef}
\begin{aligned}
\min \, &D_{KL}\left[ q(\bm{w})\,  || \, p(\bm{w}|\bm{D}) \right]=\mathbb{E}_{\bm{w}\sim q(\bm{w})}\left[ln \,q(\bm{w})-ln \,p(\bm{w}|\bm{D}) \right],
\end{aligned}
\end{equation}
Now, if we take $q(\bm{w})$ as a mixture of isotropic Gaussians, $q(\bm{w})=\frac{1}{K}\sum_{k=1}^K\mathcal{N}\left(\bm{w}_k,\sigma^2I\right)$, where $\bm{w}_k$'s are the  variational parameters to be optimized. 
When $\bm{w}_k$'s are well separated, $\mathbb{E}_{\bm{w}\sim q(\bm{w})}\left[ln \,q(\bm{w})\right]$ does not depend on $\bm{w}_i$. In the limit $\sigma\to 0$, $q(\bm{w})$ becomes the linear combination of $K$ Dirac-delta functions $\frac{1}{K}\sum_{k=1}^K\delta(\bm{w_i})$, then $\mathbb{E}_{\bm{w}\sim q(\bm{w})}\left[ln \,p(\bm{w}|\bm{D})\right]=\frac{1}{K}\sum_{k=1}^Kln \,p(\bm{w}_k|\bm{D})$. Finally, notice that data on different clients are usually independent, therefore the log posterior density factoririzes as: $
ln \, p(\bm{w}_k| \bm{D})= ln \, p(\bm{D}|\bm{w}_k) + ln \, p(\bm{w}_k)=ln \, p(\bm{w}_k)+\sum_{i}ln \, p\left(\bm{D_i}|\bm{w}\right)
$. If we take the loss function in \eqref{eqn::objective} to be $\ell_i(\bm{w}_k)=\frac{1}{n_i}ln \, p(\bm{D}_i|\bm{w}_k)+\frac{1}{\sum_in_i}p(\bm{w}_k)$, we then recover \eqref{eqn::objective}. This Bayesian view highlights the ensemble diversity as the $K$ modes can be viewed as modes of a mixture Gaussian. Further details on this Bayesian interpretation can be found in the Appendix.

\section{Convergence and limiting behavior of sufficiently over parameterized neural networks} \label{sec:theory}
In this section we present theoretical results on the generalization advantages of ensembling. We show that one can improve generalization performance by reducing predictive variance. Then we analyze sufficiently overparametrized networks trained by \textit{\name}. \textbf{We  prove the convergence of the training loss, and derive the limiting model after sufficient training, from which we show how generalization can be improved using \textit{\name}}. 

\subsection{Variance-bias decomposition} We begin by briefly reviewing the bias-variance decomposition of a general regression problem. We use $\bm{\theta}$ to parametrize the hypothesis class $h_{\bm{\theta}}$, $\overline{h_{\bm{\theta}}}(x)$ to denote the average of $h_{\bm{\theta}}(x)$ under some distribution $q(\bm{\theta})$, i.e. $\overline{h_{\bm{\theta}}}(x) = \mathbb{E}_{\bm{\theta}\sim q(\bm{\theta})}\left[h_{\bm{\theta}}(x)\right]$. Similarly $\overline{y}(x)$ is defined as $\overline{y}(x)=\mathbb{E}_{y}\left[y|x\right]$, then:
\begin{equation}
\label{eqn::vardecomposition}
\begin{array}{l}
\mathbb{E}_{y,x,\bm{\theta}}\left[\left(y-h_{\bm{\theta}}(x)\right)^2\right]=\mathbb{E}_{y,x,\bm{\theta}}\left[\left(y-\overline{y}(x)\right)^2\right]+\mathbb{E}_{x,\bm{\theta}}\left[\left(\overline{y}(x)-\overline{h_{\bm{\theta}}}(x)\right)^2\right]+\\ \mathbb{E}_{x,\bm{\theta}}\left[\left(\overline{h_{\bm{\theta}}}(x)-h_{\bm{\theta}}(x)\right)^2\right]\\
\end{array}
\end{equation}
where the expectation over $x$ is taken under some input distribution $p(x)$, and that over $\bm{\theta}$ is taken under $q(\theta)$.  In \eqref{eqn::vardecomposition}, the first term represents the intrinsic noise in data, referred to as data uncertainty. The second term is the bias term which represents the difference between expected  predictions and the expected predicted variable $y$. The third term characterizes the variance from the discrepancies of different functions in the hypothesis class, also referred to as knowledge uncertainty in \cite{gradientboosting}. In FL, this variance is often large due to heterogeneity, partial participation, etc. However, we will show that this variance decreases through model ensembling. 

\subsection{Convergence and variance reduction using neural tangent kernels}
Inspired by recent work on neural tangent kernels \cite{ntk,ntkgaussian}, we analyze sufficiently overparametrized neural networks. We focus on regression tasks and define the local empirical loss in \eqref{eqn::objective} as:

\begin{equation}
\label{eqn::squareloss}
 \ell_i(\bm{w}_k)=\frac{1}{2n_i}\sum_{(x_{ij},y_{ij})\in\bm{D}_i}\left(f_{\bm{w}_k}(x_{ij})-y_{ij}\right)^2   
\end{equation}
For this task, we will prove that when overparameterized neural networks are trained by \textit{\name}, the training loss converges exponentially to a small value determined by stepsize, and $f_{\bm{w}_k}(x)$ for all $k\in[1,.., K]$ converge to samples from the same posterior Gaussian Process ($\mathcal{GP}$) defined via a neural tangent kernel. Note that for space limitations we only provide an informal statement of the theorems, while details are relegated to the Appendix. Prior to stating our result we introduce some needed notations.  

For notational simplicity we drop the subscript $k$ in $\bm{w}_k$, and use $\bm{w}$ instead, unless stated otherwise. We let $\bm{w}(0)$ denote the initial value of $\bm{w}$, and $\bm{w}(t)$ denote $\bm{w}$ after $t$ local epochs of training. We also let $p_{\text{init}}$ denote the initialization distribution for weights $\bm{w}$. Conditions for $p_{\text{init}}$ are found in the Appendix. We define an initialization covariance matrix for any input $x$ as $\mathcal{K}(x,x^\prime)=\mathbb{E}_{\bm{w}\sim p_{\text{init}}(\bm{w})}[f_{\bm{w}}(x)f_{\bm{w}}(x^\prime)]$. Also we denote the neural tangent kernel of a neural network to be $\Theta^{(l)}(\cdot,\star)=\sum_{u=1}^{U} \partial_{w_{u}} f_{\bm{w}}(\cdot)\partial_{w_{u}} f_{\bm{w}}(\star)$, where $l$ represents the minimum width of neural network $f_{\bm{w}}$ in each layer and $U$ denotes the number of trainable parameters. Indeed, \cite{ntk} shows that this limiting kernel $\Theta$ remains fixed during training if the width of every layer of a neural network approaches infinity and when the stepsize $\eta$ scales with $l^{-1}$: $\eta=\frac{\eta_0}{l}$. We adopt the notation in \cite{ntk} and extend the analysis to FL settings. 

Below we state an informal statement of our convergence result.


\begin{theorem}
\label{thm::traininglossconverge}
(Informal) For the least square regression task $\min \frac{1}{2}\mathbb{E}_{(x,y)\sim \mathbf{D}}\left|f(x)-y\right|^2$, where $f(x)$ is a neural network whose width l goes to infinity, $l\to\infty$, then under the following assumptions 
\begin{enumerate}[label=(\roman*)]
    \item $\Theta$ is full rank i.e., $\lambda_{min}(\Theta)>0$
    \item The norm of every input x is smaller than 1: $||x||\le 1$
    \item  The stationary points of all local losses coincide: $\nabla_{\bm{w}}\mathbb{E}_{(x,y)\sim \mathbf{D}}\left|f_{\bm{w}}(x)-y\right|^2=0$ leads to \\
    $\nabla_{\bm{w}}\mathbb{E}_{(x,y)\sim \mathbf{D}_i}\left|f_{\bm{w}}(x)-y\right|^2=0$ \, for all clients
    \item The total number of data points in one communication round is a constant, $\overline{n}$
\end{enumerate}
when we use \name and local clients train via gradient descent with stepsize $\eta=\frac{\eta_0}{l}$, the training error associated to each mode decreases exponentially
$$
\mathbb{E}_{(x,y)\sim \mathbf{D}}\left|f_{\bm{w}(t)}(x)-y\right|^2\le e^{-\frac{\eta_0\lambda_{min}(\Theta) t}{3\overline{n}}}\mathbb{E}_{(x,y)\sim \mathbf{D}}\left|f_{\bm{w}(0)}(x)-y\right|^2+o(\eta_0^2)
$$
if the learning rate $\eta$ is smaller than a threshold.
\end{theorem}

\begin{remark}
Assumptions (i) and (ii) are standard for theoretical development in NTK. Assumption (iii) can be derived directly from $B$-local dissimilarity condition in \cite{fedprox}. It is actually an overparametrization condition: it says that if the gradient of the loss evaluated on the entire union dataset is zero, the gradient of the loss evaluated on each local dataset is also zero. Here we note that recent work \cite{fedpd,feddyn} have tried to provide FL algorithms that work well when this assumption does not hold. As aforementioned, such algorithms can be utilized within our ensembling framework. Assumption (iv) is added only for simplicity: it can be removed if we choose a stepsize according to number of datapoints in each round. 
\end{remark}

We would like to note that after writing this paper, we find that \cite{flntk} show convergence of the training loss in FL under a kernel regime. Their analysis, however, is limited to a special form of 2-layer \textit{relu} activated networks with the top layer fixed, while we study general multi-layer networks. Also this work is mainly concerned with theoretical understanding of FL under NTK regimes, while our overarching goal is to propose an algorithm aimed at ensembling, \textit{\name}, and motivate its use through NTK.

More importantly and beyond convergence of the training loss, we can analytically calculate the limiting solution of sufficiently overparametrized neural networks. The following theorem shows that models in the ensemble will converge into independent samples in a Gaussian Process: 
\begin{theorem}
\label{thm::ntkgaussian}
(Informal) If \name in Algorithm \ref{alg::bfl} is used to train $\{\bm{w}_k\}_{k=1,...K}$ for the regression task \eqref{eqn::objective} and \eqref{eqn::squareloss}, then after sufficient communication rounds, functions $f_{\bm{w}_k}(x)$ can be regarded as independent samples from a $\mathcal{GP}\left(m(x),k(x,x^{\prime})\right)+o(\eta_0^2)$, with mean and  variance defined as $$
m(x)=\Theta(x,X)\Theta^{-1}(X,X)Y $$,
and
$$k(x,x^\prime)=\mathcal{K}\left(x,x^\prime\right) +\Theta\left(x, X\right) \Theta^{-1} \mathcal{K}\Theta^{-1} \Theta\left(X, x^\prime\right)-\left(\Theta\left(x, X\right) \Theta^{-1} \mathcal{K}\left(X, x^\prime\right)+\Theta\left(x^\prime, X\right) \Theta^{-1} \mathcal{K}\left(X, x\right)\right)
$$,
and $(X,Y)$ represents the entire dataset $\bm{D}$.
\end{theorem} 

\begin{figure}[htbp]
\centering
\includegraphics[scale=0.5]{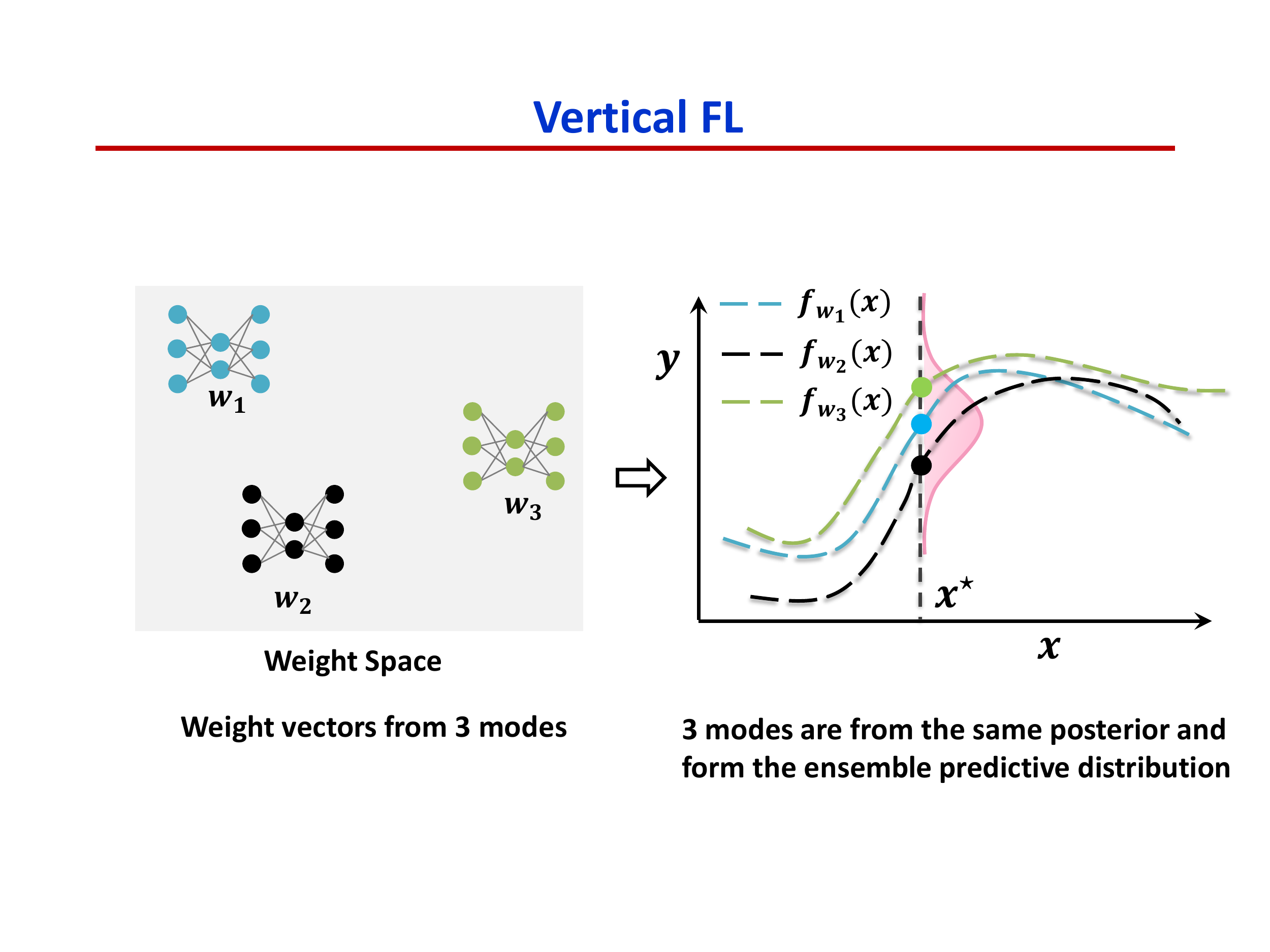}
\caption{An illustration of an ensemble of 3 samples from the posterior distribution.}
\label{fig::illustration}  
\end{figure}

\begin{remark}
The result in Theorem \ref{thm::ntkgaussian} is illustrated in Fig. \ref{fig::illustration}. The central result is that training $K$ modes with \textit{\name} will lead to predictions $f_{\bm{w}_k}(x)$, all of which are samples from the same posterior $\mathcal{GP}$. The mean of this $\mathcal{GP}$ is the exact result of kernel regression using $\Theta$, while the variance is dependent on initialization. Hence via \textit{\name} one is able to obtain multiple samples from the posterior predictive distribution. This result is similar to the simple sample-then-optimize method by \cite{matthews2017sample} which shows that for a Bayesian linear model the action of sampling from the prior followed by deterministic gradient descent  of the squared loss provides posterior samples given normal priors.
\end{remark}

A direct consequence of the result above in given in the corollary below. 

\begin{corollary}
\label{thm::illustration}
Let $\epsilon>0$ be a positive constant, if the assumptions in theorem \ref{thm::traininglossconverge} are satisfied, when we train with \name with $K$ modes, after sufficient iterations, we have that:
\begin{equation}
\label{eqn::chebyshev}
\mathbb{P}\left(\left|\frac{1}{K}\sum_{k=1}^Kf_{\bm{w}_k}-\tilde{f}\right|\ge \epsilon\right)\le O\left(\frac{1}{\epsilon^2 K}\right) 
\end{equation}
\end{corollary}
where $\tilde{f}$ is the \textit{maximum a posteriori} solution of the limiting Gaussain Process obtained in Theorem \ref{thm::ntkgaussian}. This corollary is obtained by simply combining Chebyshev inequality with Theorem \ref{thm::ntkgaussian} and shows that since the variance shrinks at the rate of $\frac{1}{K}$, averaging over multiple models is capable of getting closer to $m(x)$. 

We finally should note that there is a gap between neural tangent kernel and the actual neural network \cite{cntk}. Yet this analysis still serves to highlight the generalization advantages of FL inference with multiple modes. Also, experiments show \name works well beyond the kernel regime assumed in Theorem \ref{thm::traininglossconverge}.

\section{Experiments} 
\label{sec:numerical}

In this section we provide empirical evaluations of \textit{\name} on five representative datasets of varying sizes. We start with a toy example to explain the bias-variance decomposition, then move to realistic tasks.
We note that since ensembling is an approach yet to be fully investigated in FL, we dedicate many experiments to a proof of concept.  

\paragraph{A simple toy example with kernels}
\label{sec::toyregression}
We start with a toy example on kernel methods that illustrates the benefits of using multiple modes and reinforces the key conclusions from Theorem \ref{thm::ntkgaussian}. We create 50 clients and generate the data of each client following a noisy sine function $y=a\sin (2\pi x)+\epsilon$, where $\epsilon\sim \mathcal{N}(0, 0.2^2)$, and $a$ denotes the  parameter unique for each client that is sampled from a random distribution $\mathcal{N}(1,0.2^2)$. On each client we sample 2 points of $x$ uniformly in $[-1,1]$. We 
use the linear model $f(x;\mathcal{W})=\sum_{k=1}^{100} w_k\varphi_k(x)$, where $\varphi_k(x)$ is a radial basis kernel defined as:
$
\phi_k(x)=\exp \left(-\frac{\left(x-\nu_k\right)^2}{2b^2}\right)
$. In this badis function, $b=0.08$ and $\nu_k$'s are 100 uniformly randomly sampled parameters from $[-1,1]$ that remain constant during training. Note that the expectation of the generated function is $\mathbb{E}(y|x)=\sin (2\pi x)$. 

\begin{figure}[htbp]
\vspace{-0.5cm}
\centering   
\subfigure[Prediction performance of \name.] 
{
	\begin{minipage}{6cm}
	\centering         
	\includegraphics[scale=0.4]{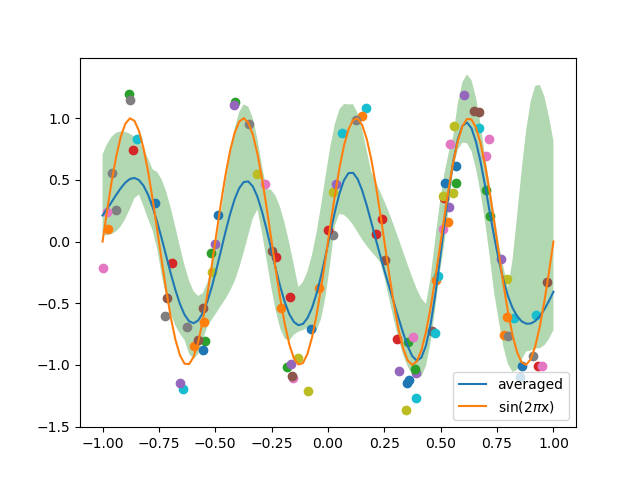}  
	\end{minipage}
}\subfigure[Prediction performance of Fedavg. ]
{
	\begin{minipage}{6cm}
	\centering      
	\includegraphics[scale=0.4]{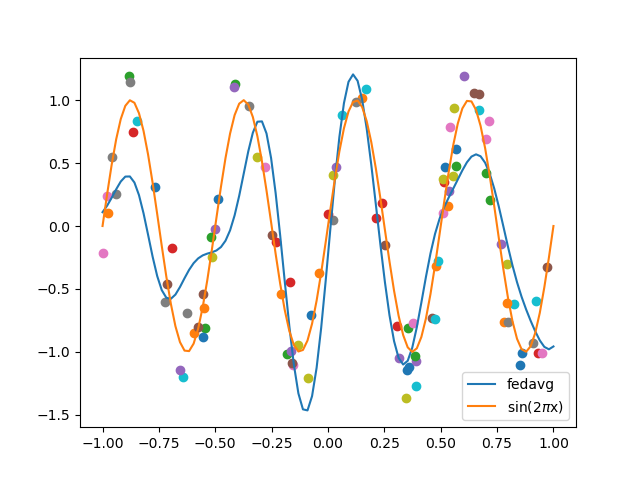}  
	\end{minipage}
}
\caption{Linear model on toy dataset. Dots represent datapoint from a subset of clients. The green area denotes predictive interval obtained from individual models, $f_{\bm{w}_k}(x)$, estimated using \textit{\name}. The ``averaged" reports the final prediction obtained after model averaging in \textit{\name}.}
\vspace{-0.1cm}
\label{fig::toymodel}  
\end{figure}

We report our predictive results in Table \ref{tab::toymodel} and Fig. \ref{fig::toymodel}. Despite the diversity of individual mode predictions (green area), upon averaging, the averaged result becomes more accurate than a fully trained single mode (Fig. \ref{fig::toymodel} (a,b)). Hence, \textit{\name} is able to average out the noise introduced by individual modes. As a more quantitative measurement, in Table \ref{tab::toymodel} we vary $K$ and calculate the bias-variance decomposition in (\ref{eqn::vardecomposition}) in each case. As seen in the table, the variance of \textit{\name} can be efficiently reduced compared with a single mode approach such as FedAvg.

\begin{table}
\begin{center}
\begin{small}
\begin{sc}
\begin{tabular}{cccccc}
\toprule
   &     K=1(FedAvg) & K=2 & K=10 & K=20 & K=40 \\
\midrule
      Bias &      0.109 &  0.117&    0.112 &      0.113 & 0.112\\

  Variance &      0.0496 &  0.0115 &   0.0063 &     0.0045 & 0.0042 \\
\bottomrule
\end{tabular}  
\end{sc}
\end{small}
\end{center}
\caption{Bias-variance decomposition on the toy regression model. We fix training data, and run each algorithm 100 times from random initializations. We take the average of 100 models as $\overline{h}_{\bm{\theta}}$ in \eqref{eqn::vardecomposition}. Bias and variance are calculated accordingly. When we increase $K$ from 1 to 40, bias almost remains in the same level, while variance decays at a rate slightly slower than $\frac{1}{K}$. Variance ceases to decrease after $K$ is larger than 20. This is intuitively understandable as when variance is very low, a larger dataset and number of communication rounds are needed to decrease it further.}
\label{tab::toymodel}
\end{table}


\paragraph{Experimental setup} 
\label{sec::setup}
In our evaluation, we show that \textit{\name} outperforms its counterparts across two popular but different settings: (i) \emph{Homogeneous setting}: data are randomly shuffled and then uniformly randomly distributed to clients; (ii) \emph{Heterogeneous setting}: data are distributed in a non-i.i.d manner: we firstly sort the images by label and then assign them to different clients to make sure that each client has images from exactly 2 labels. We experiment with  five different datasets of varying sizes: 
\begin{denseitemize}
    \item \emph{MNIST}: a popular image classification dataset with 55k training samples across 10 categories.
    \item \emph{CIFAR10}: a dataset containing 60k images from 10 classes.
    \item \emph{CIFAR100}: a dataset with the similar images of CIFAR10 but categorized into 100 classes. 
    \item \emph{Shakespeare}: the complete text of William Shakespeare with 3M characters for next work prediction.
    \item \emph{OpenImage}: a real-world image dataset with 1.1M images of 600 classes from 13k image uploaders \cite{kuznetsova2020open}. We use the realistic distribution of client data in FedScale benchmark~\cite{fedscale}. 
\end{denseitemize}
In our experiments, we use $K=5$ modes by default, except in the sensitivity analysis, where we vary $K$. Initial learning rates for clients are chosen based on the existing literature for each method. 


We compare our model with \textit{Fedavg, Fedprox, Fedbe} and \textit{Fedbe} without knowledge distillation, where we use random sampling to replace knowledge distillation. Some entries in Fedbd/Fedbe-noKD columns are missing either because it's impractical (dataset on the central server does not exist) or we cannot finetune the hyper-parameters to achieve reasonable performance. 

\begin{figure}[htbp]
\vspace{-0.2cm}
\centering
\includegraphics[scale=0.45]{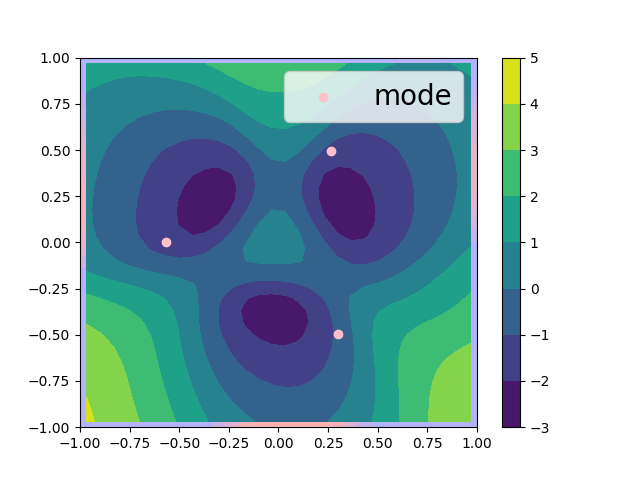}
\caption{Projection of loss surface to a plane spanned by 3 modes of the ensemble trained by \textit{\name}. Color represents the logarithm of cross-entropy loss on the entire training set.}
\label{fig::losssurface}  
\end{figure}

\paragraph{MNIST dataset.} We train a 2-layer convolutional neural network to classify different labels. Results in Table  \ref{tab::overallperformance} show that \textit{\name} outperforms all benchmarked single mode FL algorithms. This confirms the effectiveness of ensembling over single mode methods in improving generalization. Fedbe turns out to perform closely to single mode algorithms eventually. We conjecture that this happens as the diversity of local models is lost in the knowledge distillation step. To show diversity of models in the ensemble, we plot the the projection of the loss surface on a plane spanned by 3 modes from the ensemble at the end of training in Fig \ref{fig::losssurface}. Fig \ref{fig::losssurface} shows that modes ($\bm{w}_k$) have rich diversity: different modes correspond to different local minimum with a high loss barrier on the line connecting them.
In the non-i.i.d setting, the performance gap between \textit{\name} and single mode algorithms becomes larger compared with i.i.d settings. This shows that \textit{\name} can better fit more variant data distributions compared with Fedavg and Fedprox. This result is indeed expected as ensembling excels in stabilizing predictions \cite{multiswag}.

\paragraph{CIFAR/Shakespeare/OpenImage dataset.} We use ResNet \cite{resnet} for CIFAR, a character-level LSTM model for the sequence prediction task, and ShuffleNet \cite{shufflenet} and MobileNet \cite{mobilenet} for OpenImage.  Results in Table \ref{tab::overallperformance} show that \textit{\name} can indeed improve the generalization performance. Further details on the experimental setup can be found in the Appendix.

\begin{table}
\begin{center}
\begin{small}
\begin{sc}

\begin{tabular}{cccccc}
\toprule
   Testing acc (\%) &     Fedavg &    Fedprox &      Fedbe & Fedbe-nokd &      \name \\
\midrule
 MNIST-iid &       $95.08\pm 0.3$ &      $95.76\pm0.12$ &      $95.21\pm 0.11$ &          - &      \textbf{97.75$\pm 0.04$} \\

MNIST-noniid &       $90.17\pm 0.19$ &      $90.75\pm0.34$ &          - &          - &      \textbf{95.44$\pm0.14$} \\

CIFAR10-iid &      $86.43\pm0.11$ &      $86.77\pm0.06$ &      $83.13\pm0.34$ &      $86.78\pm0.4$ &      \textbf{87.99$\pm 0.09$} \\

CIFAR10-noniid &      $66.11\pm0.17$ &      $66.62\pm 0.24$ &          - &          - &      \textbf{71.18$\pm0.4$} \\

  CIFAR100 &      $56.25\pm0.22$ &      $55.91\pm0.25$ &      $49.24\pm 2$ &          - &      \textbf{58.09$\pm0.12$} \\

Shakespeare &      $60.05\pm0.02$ &      $60.18\pm0.02$ &          - &      $61.52\pm0.22$ &       \textbf{62.49$\pm0.08$} \\

OpenImage mobile net &      51.85$\pm$0.17 &      52.93$\pm$0.14 &          - &          - &      \textbf{53.92}$\pm$0.19 \\

OpenImage shuffle net &      53.98$\pm$0.16 &      54.42$\pm$0.22 &          - &          - &      \textbf{55.75}$\pm$0.25 \\

\bottomrule
\end{tabular}  
\end{sc}
\end{small}
\end{center}
\caption{Testing accuracy of models trained by different FL algorithms on five datasets. Fedbe-Nokd denotes Fedbe without knowledge distillation.}
\label{tab::overallperformance}
\end{table}

\paragraph{Effect of non-i.i.d client data distribution}
\label{sec:ablation}

\begin{table}
\begin{center}
\begin{small}
\begin{sc}
\begin{tabular}{lcccr}
\toprule
N.o.C. & Fedavg & Fedprox & \name & Gap \\
\midrule
2   & 66.19 & 66.87 & \textbf{71.45} & 4.58\\
4   & 83.90& 84.40 & \textbf{86.12} & 1.72\\
6    & 85.90 & 86.10 & \textbf{87.14} & 1.04\\
8    & 85.90& 86.06 & \textbf{87.33} & 1.27\\
10    & 86.52 & 86.77 & \textbf{87.94} & 1.17\\

\bottomrule
\end{tabular}
\end{sc}
\end{small}
\end{center}
\caption{Sensitivity analysis with different assigned N.o.C on CIFAR-10. Gap denotes the difference in testing accuracy between \textit{\name} and Fedprox.}
\label{tab::noc}
\vspace{-1em}
\end{table}

We change the number of classes (N.o.C.) assigned to each client from 10 to 2 on the CIFAR-10 dataset. Conceivably, when each client has fewer categories, the data distribution is more variant. The results are shown in Table \ref{tab::noc}. As expected, the performance of all algorithms degrades with such heterogeneity. Further \textit{\name} outperforms its counterparts and the performance gap becomes significantly clear as variance increases (i.e., smaller N.o.C.). This highlights the ability of \textit{\name} to improve generalization specifically with heterogeneous clients.

\paragraph{Effect of number of modes $K$.}
\label{sec::sensitivity}
Since the number of modes, $K$, in the ensemble is an important hyperparameter, we choose different values of $K$ and test the performance on MNIST. We vary $K$ from 3 to 80. The results are shown in Table \ref{tab::ksensitivity}. Besides testing accuracy of ensemble predictions, we also calculate the accuracy and the entropy of the predictive distribution of each mode in the ensemble.
As shown in Table \ref{tab::ksensitivity}, as $K$ increases from 3 to 40, the ensemble prediction accuracy increases as a result of variance reduction. However, when $K$ is very large, entropy increases and model accuracy drops slightly, suggesting that model prediction is less certain and accurate. The reason here is that when $K=80$, the number of clients, hence datapoints, assigned to each mode significantly drop. This in turn decreases learning accuracy specifically when datasets are relatively small and a limited budget exists for communication rounds. 

\begin{table}[h]
\begin{center}
\begin{small}
\begin{sc}
\begin{tabular}{cccccc}
\toprule
           &        K=3 &        K=5 &       K=20 &       K=40 &       K=80 \\
\midrule
  Test acc & $97.49\pm 0.2$ & $97.85\pm 0.02$ & $98.19\pm 0.08$ & $98.17\pm 0.01$ & $97.98\pm 0.03$ \\

Acc max &       95.50 &       95.60 &      95.57 &      95.53 &      94.84 \\

Acc min &      94.97 &      95.00 &      94.13 &      93.54 &      92.74 \\

Avg entropy &     0.16 &     0.15 &     0.16 &     0.18 &     0.20 \\

\bottomrule
\end{tabular}  
\end{sc}
\end{small}
\end{center}
\caption{Sensitivity of number of modes $K$ on MNIST. Acc max/min is the maximum/minimum testing accuracy across all modes in the ensemble. Avg Entropy is the average of the entropy of the empirical predictive distribution across all modes.}
\label{tab::ksensitivity}
\vspace{-1em}
\end{table}
\section{Conclusion \& potential extensions}
This paper proposes \textit{\name}: an ensembling approach for FL which learns $K$ modes and obtains predictions via model averaging. We show, both theoretically and empirically, that \textit{\name} is efficient in reducing variance and achieving better generalization performance in FL compared to single mode approaches. Beyond ensembling, \textit{\name} may find value in meta-learning where modes acts as multiple learned initializations and clients download ones with optimal performance based on some training performance metric. This may be a route worthy of investigation. Another potential extension is instead of using random permutations, one may send specific modes to individual clients based on training loss or gradient norm, etc., from previous communication rounds. The idea is to increase assignments for modes with least performance or assign such modes to
clients were they mal-performed in an attempt to improve the worst performance cases. Here it is interesting to understand if theoretical guarantees still hold in such settings.

\newpage
\bibliography{bfl.bib}
\bibliographystyle{plain}

\newpage
\section{Additional Experimental Details}
In this section we articulate experimental details, and present round-to-accuracy figures to better visualize the training process. In all experiments, we decay learning rates by a constant of 0.99 after 10 communication rounds. On image datasets, we use stochastic gradient descent (SGD) as the client optimizer, while on the Shakespear dataset, we train using Adam. Every set of experiment can be done on single Tesla V100 GPU.

\begin{figure}[htbp]
\vspace{-0.5cm}
\centering   
\subfigure[MNIST iid.] 
{
	\begin{minipage}{6cm}
	\centering         
	\includegraphics[scale=0.4]{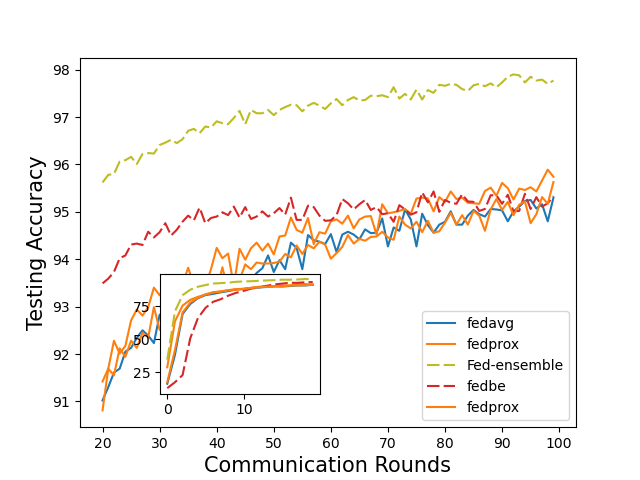}  
	\end{minipage}
}\subfigure[MNIST noniid. ]
{
	\begin{minipage}{6cm}
	\centering      
	\includegraphics[scale=0.4]{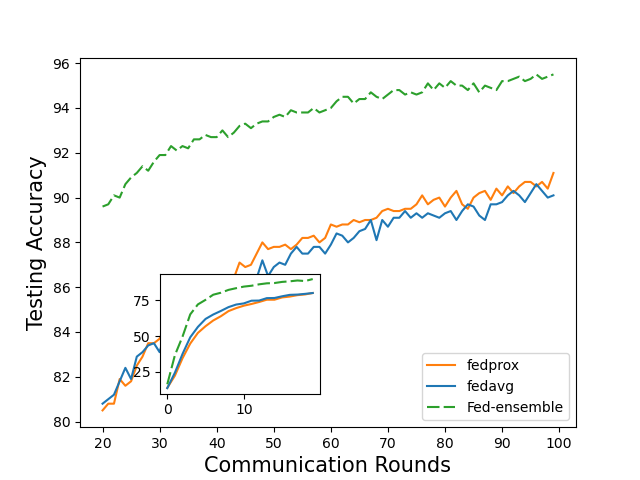}  
	\end{minipage}
}
\caption{Training process on MNIST}
\vspace{-0.1cm}
\label{fig::mnist}  
\end{figure}

Fig \ref{fig::mnist} is the round-to-accuracy of several algorithms on MNIST dataset. \name  outperforms all single mode federated algorithms and Fedbe.

Fig \ref{fig::cifar10} shows the result on CIFAR10. \name has better generalization performance from the very beginning.
\begin{figure}[htbp]
\vspace{-0.5cm}
\centering   
\subfigure[CIFAR10 iid.] 
{
	\begin{minipage}{6cm}
	\centering         
	\includegraphics[scale=0.4]{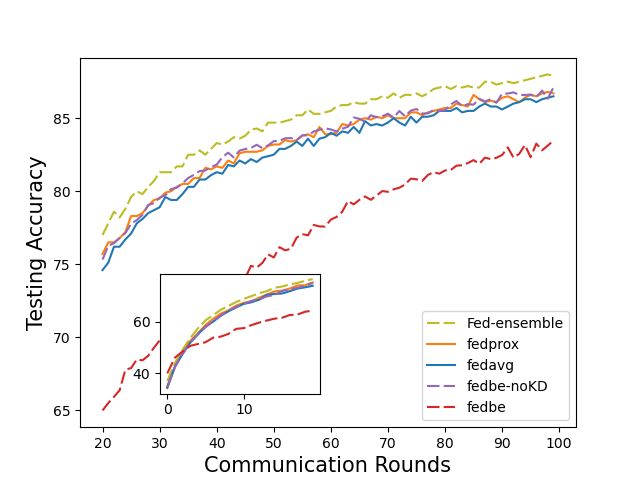}  
	\end{minipage}
}\subfigure[CIFAR10 noniid. ]
{
	\begin{minipage}{6cm}
	\centering      
	\includegraphics[scale=0.4]{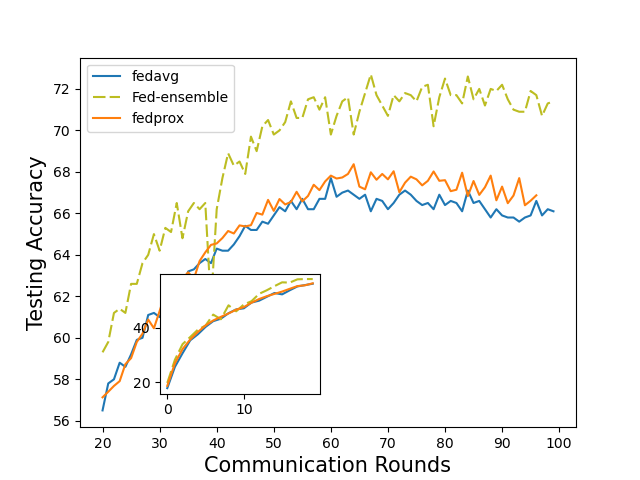}  
	\end{minipage}
}
\caption{Training process on CIFAR10}
\vspace{-0.1cm}
\label{fig::cifar10}  
\end{figure}

\begin{figure}[htbp]
\vspace{-0.5cm}
\centering   
\subfigure[Training process on CIFAR100] 
{
	\begin{minipage}{6cm}
	\centering         
	\includegraphics[scale=0.4]{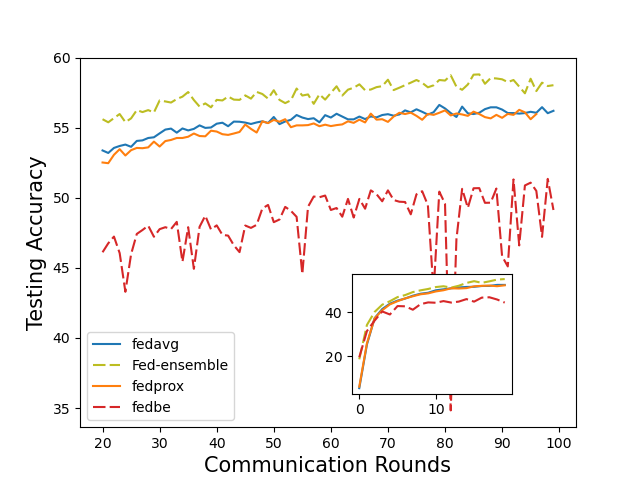}  
	\label{fig::cifar100}  
	\end{minipage}
}\subfigure[Training process on Shakespeare ]
{
	\begin{minipage}{6cm}
	\centering      
	\includegraphics[scale=0.4]{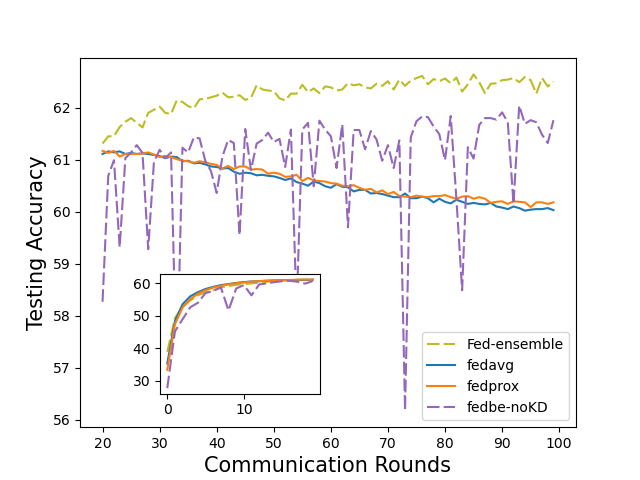}  
	\label{fig::shakespeare}  

	\end{minipage}
}
\caption{Larger datasets}
\vspace{-0.1cm}
\label{fig::cifar10}  
\end{figure}

Fig \ref{fig::cifar100} plots round-to-accuracy on CIFAR100. Also, there is a clear gap between \name and FedAvg/FedProx.

The comparison between \name and single mode algorithms is even more conspicuous on Shakespeare. The next-word-prediction model we use has 2 LSTM layers, each with 256 hidden states. Fig \ref{fig::shakespeare} shows that FedAvg and FedProx suffers from severe overfitting when communication rounds exceed 20, while \name continues to improve. Fedbe is unstable on this dataset.


On OpenImage dataset, we simulate the real FL deployment in Google \cite{google-fl}, where we select 130 clients to participate in training, but only the updates from the first 100 completed clients are aggregated in order to mitigate stragglers.

\section{Additional Details on Deriving the Variational Objective}
As discussed in Sec. \ref{sec::bayesianinterpretation}
the objective is to minimize the KL-divergence between the variational and true posterior distributions. \begin{equation}
\label{eqn::kldivdef}
\begin{aligned}
\min \, &D_{KL}\left[ q(\bm{w})\,  || \, p(\bm{w}|\bm{D}) \right]\\
&=\mathbb{E}_{\bm{w}\sim q(\bm{w})}\left[ln \,q(\bm{w})-ln \,p(\bm{w}|\bm{D}) \right],
\end{aligned}
\end{equation}

In this section, we use Gaussian mixture as variational distribution and then take limit $\sigma\to 0$ to the result. Given (\ref{eqn::kldivdef}), $D_{KL}\left[ q(\bm{w})\,  || \, p(\bm{w}|\bm{D}) \right]$ can be decomposed by conditional expectation law:
\begin{align}
\mathbb{E}_{k\sim Uniform(K)}\left[\mathbb{E}_{\bm{w}\sim \mathcal{N}(\bm{w}_k, \sigma^2\bm{I})}\left[ln \,q(\bm{w})-ln \,p(\bm{D}|\bm{w})\right] \right], \notag
\end{align}
where \textit{Uniform(K)} denotes a uniform distribution on $[K]$. When $\sigma$ is very small, different mode centers ($\bm{w}_k$) are well-separated such that $q(\bm{w})$ is close to zero outside a small vicinity of $\bm{w}_k$. Thus,
\begin{align}
&\mathbb{E}_{\bm{w}\sim \mathcal{N}(\bm{w}_k, \sigma^2\bm{I})}\left[\ln \,q(\bm{w})\right] \notag\\
&\approx \mathbb{E}_{\bm{w}\sim \mathcal{N}(\bm{w}_k, \sigma^2\bm{I})}\left[-\frac{1}{2\sigma^2}\left\|\bm{w}-\bm{w}_k\right\|^2-d\ln\sigma\right]+C \notag\\
&=-\frac{1}{2}-d\ln\ \sigma + C
\end{align}
where $C$ is a constant. Thus this term is independent of $k$. Now taking a first order Taylor expansion of
$ln \,p(\bm{w}|\bm{D})$ around $\bm{w}_k$, we have
\begin{align}
ln \,p(\bm{w}|\bm{D})& \approx ln \,p(\bm{w}_k|\bm{D})+\left(\bm{w}-\bm{w}_k\right)^T\nabla ln \,p(\bm{w}_k|\bm{D}). \notag
\end{align}
After taking expectation on $\bm{w}$, we have $\mathbb{E}_{\bm{w}\sim \mathcal{N}(\bm{w}_k,  \sigma^2\bm{I})}\left[ln\,p(\bm{w}|\bm{D})\right]\approx ln\,p(\bm{w}_k|\bm{D})$ in the small $\sigma$ limit. As a result, by conditional expectation, the KL-divergence reduces to:
\begin{align} \label{eqn::kldiv-objective}
&D_{KL}\left(q(\bm{w}) \, || \, p(\bm{w}|\bm{D}) \right) \notag \\
&= \mathbb{E}_{k\sim Uniform(K)}\left[-ln\,p(\bm{w}_k|\bm{D}) \right]+C \notag\\
&=\frac{1}{K}\sum_{k=1}^K-\ln\,p(\bm{w}_k)-\ln\, p(\bm{D}|\bm{w}_k)+C^\prime.
\end{align}

where $p(\bm{D}|\bm{w}_k)$ is a function of $\bm{w}_k$ and $p(\bm{w}_k)$ is the prior p.d.f. of $p(\bm{w})$ when $\bm{w}=\bm{w}_k$.

In \eqref{eqn::kldiv-objective}, $p(\bm{w}_k)$ acts as a regularizer on negative log-likelihood loss $p(\bm{D}|\bm{w}_k)$. When the prior is Gaussian centered at $0$, the prior term simply becomes $L_2$ regularization.

	\section{Proof of theorem \ref{thm::traininglossconverge} and \ref{thm::ntkgaussian}}
	In section, we discuss the training of linear models and neural networks in the sufficiently overparametrized regime, and prove theorem \ref{thm::traininglossconverge} and \ref{thm::ntkgaussian}. This section is organized as following: in sec \ref{ap::notations}, we introduce some needed notations and prove some helper lemmas. In sec \ref{ap::trainlinearmodel}, we describe the dynamics of training linear model using \name.
	In sec \ref{ap::trainneuralnet}, we prove the global convergence of training loss of \name, and show that neural tangent kernel is fixed during training of \name at the limit of width $l\to\infty$. The result is a refined version of theorem \ref{thm::traininglossconverge}.  In sec \ref{sec:difbound} we will show that the training can indeed be well approximated by kernel method in the extreme wide limit. Then combining sec \ref{sec:difbound} with sec \ref{ap::trainneuralnet}, we can prove theorem \ref{thm::ntkgaussian}.
	\subsection{Notations and Some lemmas}
	\label{ap::notations}
	In this section we will introduce some notations, followed by some lemmas.
	
	For an operator $A$, we use $\left\|A\right\|_{op}$ to denote its operator norm, i.e. largest eigenvalue of $A$. If $A$ is a matrix, $\left\|A\right\|_{F}$ is its Frobenius norm. Generally we have $\left\|A\right\|_{F}>\left\|A\right\|_{op}$. $\lambda_{min}(A)$ is the smallest eigenvalue of $A$. 
	
	The update of $K$ different modes in the ensemble described by algorithm 1 in the main text is decoupled, thus we can monitor the training of each mode separately and drop the subscript $k$. We focus on one mode and use $\bm{w}$ to denote the weight vector of this mode. We define $g(\bm{w})$ to be an $n$-dimensional error vector of training data on all clients, $g(\bm{w})=f_{\bm{w}}(X)-Y$. Its $j$-th component is $f_{\bm{w}}(x_j)-y_j$. Notice that $g$ contains the datapoints from all clients. Similarly we can define $g_i$ to be the error vector of the $i$-th client $g_i(\bm{w})=f_{\bm{w}}(X_i)-Y_i$. $g(\bm{w})$ is thus the concatenation of all $g_i(\bm{w})$. Also, we can define a $n_i\times n$ projection matrix $\mathcal{P}_i$:
	\[
	\mathcal{P}_i=\begin{pmatrix}
0&\cdots&0&1 & 0&\cdots & 0 &0 &\cdots&0\\
0&\cdots&0&0 & 1&\cdots & 0 &0 &\cdots&0\\
&\vdots& & & &\ddots& & &\vdots &\\
0&\cdots&0&0 & 0&\cdots & 1 &0 &\cdots&0\\
\end{pmatrix}
	\]
	It is straightforward to verify that $g_i(\bm{w})=\mathcal{P}_ig(\bm{w})$. As introduced in the main text, we use $\bm{w}(t)$ to denote the weight vector $\bm{w}$ after $t$ iterations of local gradient descent. Suppose in each communication round, every client performs $\tau$ steps of gradient descent, then $\bm{w}(tK\tau+r\tau)$ naturally means the weight vector at the beginning of $r$-th communication round of $t$-th age. We use $J(\bm{w})$ to denote the $d\times n$ gradient matrix, whose $ij$-th element $\frac{\partial}{\partial w_i}f_{\bm{w}}(x_j)$. Similarly, $J_i(\bm{w})$ is a $d\times n_i$ gradient matrix on client $i$, whose $kj$-th element is $\frac{\partial}{\partial w_k}f_{\bm{w}}(x_j)$.
	
   We use $\eta$ to denote the constant stepsize of gradient descent in local client training. For NTK parametrization, stepsize should decrease with the increase of network width, hence we rewrite $\eta$ as $\eta=\frac{\eta_0}{l}$, where $\eta_0$ is a new constant independent of neural network width $l$. We use $\mathcal{S}\left(t,r\right)$ to denote the subset of all clients participating in training of the $r$-th communication round of the $t$-th age. We will assume that in one communication round, the number of datapoints on all clients that participated in training is the same: $\sum_{i\in\mathcal{S}\left(t,r\right)}n_i=\bar{n}, \forall r$. This assumption is roughly satisfied in our experiments. With some modifications in the weighting parameters in algorithm \ref{alg::bfl}, the equal-datapoint assumption can be alleviated.
	
   The following lemma is proved as Lemma 1 in \cite{ntkgaussian}:
	\begin{lemma}
		\label{lm::jacobianlip}
		Local Lipschitz of Jacobian. If (i) the training set is consistent: for any $x_i\neq x_{i^\prime}$, $y_i\neq y_{i^\prime}$. (ii) activation function of the neural nets $\phi$ satisfies $\phi(0)$,
		$\left\|\phi^\prime\right\|_\infty$,
		$\frac{|\phi^\prime(x)-\phi^\prime(x^\prime)|}{|x-x^\prime|}$ $<\infty$. (iii)
		$\left\|x\right\|\le 1$ for all input $x$, then there are constants
		$C^{(l)},C^{(l)}_1,...C^{(l)}_N$ such that for every $C>0$, with high probability over random initialization the following holds:
		$$
		\left\{\begin{array}{ll}
		\frac{1}{\sqrt{l}}\|J_i(\bm{w})-J_i(\tilde{\bm{w}})\|_{F} & \leq C^{(l)}_i\|\bm{w}-\tilde{\bm{w}}\|_{2} \\
		\frac{1}{\sqrt{l}}\|J_i(\bm{w})\|_{F} & \leq C_i^{(l)}\\
		\end{array} \right.
		$$
		$\forall \bm{w}, \tilde{\bm{w}} \in B\left(\bm{w}_{0}, C l^{-\frac{1}{2}}\right)$ for every i=1,2,...N, and also:
		$$
		\left\{\begin{array}{ll}
		\frac{1}{\sqrt{l}}\|J(\bm{w})-J(\tilde{\bm{w}})\|_{F} & \leq C^{(l)}\|\bm{w}-\tilde{\bm{w}}\|_{2} \\
		\frac{1}{\sqrt{l}}\|J(\bm{w})\|_{F} & \leq C^{(l)}
		\end{array} \right.
		$$
		$\forall \bm{w}, \tilde{\bm{w}} \in B\left(\bm{w}_{0}, C l^{-\frac{1}{2}}\right)$, where 
		$$
		B\left(\bm{w}_{0}, R\right):=\left\{\bm{w}:\left\|\bm{w}-\bm{w}_{0}\right\|_{2}<R\right\}
		$$
	\end{lemma}
Lemma \ref{lm::jacobianlip} discusses the sufficient conditions for the Lipschitz continuity of Jacobian. 
In the following derivations, we use the result of Lemma \ref{lm::jacobianlip} directly.

In the proof, we extensively use Taylor expansions over the product of local stepsize and local training epochs $\tau$. As we will show, the first order term can yield the desired result. We will bound the second order term and neglect higher order terms by incorporating them into an $o(\eta_0^2\tau^2)$ residual. We abuse this notation a little: $o\left(\eta_0^2\tau^2\right)$ sometimes denote a constant containing orders higher than $\eta_0^2\tau^2$, sometimes a vector whose $L_2$ norm is $o\left(\eta_0^2\tau^2\right)$, sometimes an operator whose operator norm is $o\left(\eta_0^2\tau^2\right)$. The exact meaning should be clear depending on the context. Taylor expansion is feasible when $\eta_0\tau$ is small enough. Extensive experimental results confirm the validity of Taylor expansion treatment.

	\begin{lemma}
		\label{lm::bch}
		For K operators $A_1$, $A_2$,...$A_K$, if their operator norm is bounded by $\rho_i$, $i=1\cdots K$ respectively, we have:
		\begin{equation}
		\begin{aligned}
		&e^{-\eta\tau A_K}e^{-\eta\tau A_{K-1}}...e^{-\eta\tau A_1}\\
		&=e^{-\eta\tau\left(A_K+A_{K-1}+...+A_1\right)}+\frac{1}{2}\eta^2\tau^2\sum_{i>j}\left[A_i,A_j\right]+o\left(\eta^2\tau^2\right)
		\end{aligned}
		\end{equation}
		where $[A_i,A_j]$ is defined as the commutator of operator $A_i$ and $A_j$: $[A_i,A_j]=A_iA_j-A_jA_i$. Here $o\left(\eta^2\tau^2\right)$ denotes some operator whose operator norm is $o\left(\eta\tau^2\right)$.
	\end{lemma}
	Lemma \ref{lm::bch} connects the product of exponential to the exponential of sum. This is an extended version of Baker Campbell Hausdorff formula. We still give a simple proof at the end of this section for completeness.
	
	Lemma \ref{lm::varofoperator} bounds the difference between the average of exponential and exponential of average:
	\begin{lemma}
		\label{lm::varofoperator}
		For K operators $A_1$, $A_2$,...$A_K$, and K non-negative values $p_1,...p_K$ satisfying $p_1+...+p_K=1$, we have:
		\begin{equation}
		\begin{aligned}
		&\sum_{i=1}^Kp_ie^{-\eta\tau A_i}\\
		&=e^{-\eta\tau \sum_{i=1}^Kp_iA_i}+\frac{\eta^2\tau^2}{2}\left(\sum_{i=1}^Kp_iA_i^2-\left(\sum_{i=1}^Kp_iA_i\right)^2\right)+o\left(\eta^2\tau^2\right)
		\end{aligned}
		\end{equation}
	\end{lemma}
	\begin{proof}
	The proof is straightforward by using Taylor expansion on both sides and retain only highest order terms in residual. For an operator $A$, its exponential is defined as:
	\begin{equation}
	\label{eqn::operatorexponential}
	 \begin{aligned}
	&\exp \left(-\eta\tau A\right)\\
	&=\sum_{k=0}^{\infty}\left(-1\right)^k\frac{\left(\eta\tau A\right)^k}{k!}\\
	&=1-\eta\tau A+\frac{\eta^2\tau^2}{2}A^2+\eta^3\tau^3\Delta_{A}
	\end{aligned}   
	\end{equation}
	where $\Delta_A$ is defined as:
	\begin{equation}
	\label{eqn::defdeltaa}
	 \Delta_A=\sum_{k=3}^{\infty}\left(-1\right)^k\frac{\left(\eta\tau \right)^{k-3}A^k}{k!}   
	\end{equation}

	When $A$'s operator norm is bounded by $\rho_A$, we can also bound the operator norm of $\Delta_A$ as:
	$$
	\begin{aligned}
	\left\|\Delta_A\right\|&\le\sum_{k=3}^{\infty}\frac{\left(\eta\tau \right)^{k-3}\left\|A\right\|^k}{k!}\\
	&=\left\|A\right\|^3\sum_{k=3}^{\infty}\frac{\left(\eta\tau \right)^{k-3}\left\|A\right\|^{k-3}}{k!}\\
    &\le\frac{\left\|A\right\|^3}{6}\sum_{k=3}^{\infty}\frac{\left(\eta\tau \right)^{k-3}\left\|A\right\|^{k-3}}{(k-3)!}\\
    &=\frac{\rho_A^3}{6}\exp \left(\eta\tau \rho_A\right)
	\end{aligned}
	$$
	
	Then we can rewrite two sides of the equations using $\Delta_{A_i}$ as below:
	\[
	\begin{aligned}
	&\sum_{i=1}^Kp_ie^{-\eta\tau A_i}-e^{-\eta\tau \sum_{i=1}^Kp_iA_i}\\
	&=\sum_{i=1}^Kp_i\left(1-\eta\tau A_i+\frac{1}{2}\eta^2\tau^2 A_i^2+\eta^3\tau^3\Delta_{A_i}\right)\\
	&-\left(1-\eta\tau \sum_{i=1}^Kp_iA_i+\frac{1}{2}\eta^2\tau^2 \left(\sum_{i=1}^Kp_iA_i\right)^2+\eta^3\tau^3\Delta_{\sum_{i=1}^Kp_iA_i}\right)\\
	&=\frac{1}{2}\eta^2\tau^2\left(\sum_{i=1}^Kp_iA_i^2-\left(\sum_{i=1}^Kp_iA_i\right)^2\right)+\eta^3\tau^3\left(\sum_{i=1}^Kp_i\Delta_{A_i}-\Delta_{\sum_{i=1}^Kp_iA_i}\right)
	\end{aligned}
	\]
	Since we have shown $\left\|\Delta_{A_i}\right\|$'s are bounded, the last term $\eta^3\tau^3\left(\sum_{i=1}^Kp_i\Delta_{A_i}-\Delta_{\sum_{i=1}^Kp_iA_i}\right)=O(\eta^3\tau^3)$
	\end{proof}
	Proof of Lemma \ref{lm::bch}
	\begin{proof}
	We adopt the notations in the proof of lemma \ref{lm::varofoperator} from \eqref{eqn::operatorexponential} and \eqref{eqn::defdeltaa}. Then we have:
	$$
	\begin{aligned}
	&e^{-\eta\tau A_K}e^{-\eta\tau A_{K-1}}...e^{-\eta\tau A_1}-e^{-\eta\tau\left(A_K+A_{K-1}+...+A_1\right)}\\
	&=\left(1-\eta\tau A_K+\frac{\eta^2\tau^2}{2}A_K^2+\eta^3\tau^3\Delta_{A_K}\right)\cdots\left(1-\eta\tau A_1+\frac{\eta^2\tau^2}{2}A_1^2+\eta^3\tau^3\Delta_{A_1}\right)\\
	&-\left(1-\eta\tau \left(A_K+\cdots+A_1\right)+\frac{\eta^2\tau^2}{2}\left(A_K+\cdots+A_1\right)^2+\eta^3\tau^3\Delta_{A_K+\cdots+A_1}\right)\\
	&=\frac{1}{2}\eta^2\tau^2\sum_{i>j}\left[A_i,A_j\right]+o\left(\eta^2\tau^2\right)\\
	\end{aligned}
	$$
	\end{proof}
	\subsection{Training of Linearized Neural Networks}
	\label{ap::trainlinearmodel}
	The derivation in this section is inspired by the kernel regression model from \cite{ntk}. Now $\Theta(\cdot,\star)$ can be any positive definite symmetric kernel, later we will specify it to be neural tangent kernel. We define the gram operator of client $i$ to be a map from function $f(x)$ into function
	$$
	\Pi_i(f)(x)=\frac{1}{n_i} \sum_{\left(x_{ij},y_{ij}\right)\in\bf{D}_{i}}f\left(x_{ij}\right) \Theta\left(x_{ij}, x\right)
	$$ 
	and also, the gram operator of the entire dataset is defined as:
	$$
	\Pi(f)(x)=\frac{1}{n} \sum_{\left(x_{j},y_{j}\right)\in\bf{D}}f\left(x_{j}\right) \Theta\left(x_{j}, x\right)
	$$
	In the beginning of $r$-th communication round of age $t$, $\bm{w}(tK\tau+r\tau)$ is sent to all clients in some selected strata. As discussed before, we use $S(t,r)$ to denote the set of clients that download mode $\bm{w}$ in the $r$-th communication round of age $t$. Since client $i$ uses gradient descent to minimize the objective, we approximate it with gradient flow:
	$$
	\frac{\partial}{\partial s}f_{\bm{w}^{(i)}(tK\tau+r\tau+s)}(x)=-\eta\Pi_if_{\bm{w}^{(i)}(tK\tau+r\tau+s)}(x)
	$$
	where $\bm{w}^{(i)}(t K\tau+r\tau+s)$ is the local weight vector on client $i$ starting from $\bm{w}^{(i)}(t K\tau+r\tau)=\bm{w}(t K\tau+r\tau)$.
		
	Since $\Pi_i$ does not evolve over time, after $\tau$ epochs, the updated function on client $i$ is:
	$$
	f_{\bm{w}^{(i)}(tK\tau+(r+1)\tau)}(x)-f^\star(x)=e^{-\eta\Pi_i\tau}(f_{\bm{w}^{(i)}(tK\tau+r\tau)}(x)-f^\star(x))
	$$
	where $f^\star$ is the ground truth function satisfying $\Pi_if^\star=0$ for every $i$, i.e. it achieves zero training error on every client. After $\tau$ epochs, all clients send updated model back to server, and server use $server\_update$ to calculate new $\bm{w}$. Since we are training a linear model:
	
	\begin{equation*}
	f_{\bm{w}(tK\tau+(r+1)\tau)}(x)=\frac{1}{\sum_{i\in S(t,r)}n_i}\sum_{i\in S(t,r)}n_i f_{\bm{w}^{(i)}(tK\tau+(r+1)\tau)}(x)
	\end{equation*}
	Plugging in the equation of $f_{\bm{w}^{(i)}(tK\tau+(r+1)\tau)}$, we can use lemma \ref{lm::varofoperator} to derive:
	\begin{equation*}
	\begin{aligned}
	&f_{\bm{w}(tK\tau+(r+1)\tau)}(x)-f^\star(x)\\
	&=\frac{1}{\sum_{i\in S(t,r)}n_i}\sum_{i\in S(t,r)}n_ie^{-\eta\Pi_i\tau}(f_{\bm{w}(tK\tau+r\tau)}(x)-f^\star)\\
	&=e^{-\eta\frac{1}{\sum_{i\in S(t,r)}n_i}\sum_{i\in S(t,r)}n_i\Pi_i\tau}(f_{\bm{w}(tK\tau+r\tau)}(x)-f^\star)+\\
	&\frac{1}{2}\mathbf{Var}\left(\Pi_i\right)\tau^2\eta^2(f_{\bm{w}(tK\tau+r\tau)}(x)-f^\star)+o\left(\eta^2\tau^2\right)
	\end{aligned}
	\end{equation*}
	where $\mathbf{Var}\left(\Pi_i\right)$ is defined as:
	\begin{equation*}
	\begin{aligned}
	\mathbf{Var}\left(\Pi_i\right)=\frac{1}{\sum_{i\in S(t,r)}n_i}\sum_{i\in S(t,r)}n_i\Pi_i^2-\left(\frac{1}{\sum_{i\in S(t,r)}n_i}\sum_{i\in S(k,t)}n_i\Pi_i\right)^2
	\end{aligned}
	\end{equation*}
	
	We call this term the variance of gram operator $\Pi$. When $\Pi_i$'s are very different from each other, i.e. data distribution is very heterogeneous, this term would be large and when $\Pi_i$'s are similar, this term would be small. $o\left(\eta^2\tau^2\right)$ are terms that contain $\eta\tau$ order higher than 2. When $\eta\tau$ are chosen to be not too large, we neglect higher order residuals and retain only the leading terms in further derivations. 
	
	We then consider the training of K consecutive communication rounds:
	\begin{equation}
	\label{eqn::fkupdatektaustep}
	\begin{aligned}
	&f_{\bm{w}_{(t+1)K\tau}}(x)-f^\star(x)\\
	&=\prod_{l=1}^Ke^{-\eta\mathbb{E}[\Pi_l]\tau}(f_{\bm{w}(tK\tau)}(x)-f^\star)+\\
	&\frac{1}{2}\tau^2\eta^2\sum_j\prod_{l<j}e^{-\eta\tau\mathbb{E}[\Pi_l]}\mathbf{Var}[\Pi_j]\prod_{l>j}e^{-\eta\tau\mathbb{E}[\Pi_l]}(f_{\bm{w}(tK\tau)}(x)-f^\star)+o\left(\tau^2\eta^2\right)
	\end{aligned}
	\end{equation}
	where $\mathbb{E}[\Pi_l]$ is defined as:
	\begin{equation*}
	\mathbb{E}[\Pi_l]=\frac{1}{\sum_{i\in S(t, l)}n_i}\sum_{i\in S(t,l)}n_i\Pi_i
	\end{equation*}
	
	We can reorganize operators that apply on $f_{\bm{w}}-f^\star$. We introduce a new operator $U$ as:
	\begin{equation*}
	\begin{aligned}
	U=\left(1+\frac{1}{2}\tau^2\eta^2\left(\sum_{j=1}^K\mathbf{Var}[\Pi_j]+\sum_{i> j}\left[\mathbb{E}[\Pi_i],\mathbb{E}[\Pi_j]\right]\right)\right)e^{-\eta\tau\sum_{l=1}^K\mathbb{E}[\Pi_l]}  
	\end{aligned}
	\end{equation*}
	
	By definition of matrix $\Pi$, we know that $\sum_{l=1}^K\mathbb{E}\left[\Pi_i\right]=K\Pi$, therefore, we can replace $\sum_{l=1}^K\mathbb{E}\left[\Pi_i\right]$ by $K\Pi$. By applying lemma \ref{lm::bch} and neglecting higher order terms, equation \eqref{eqn::fkupdatektaustep} can be rewritten as:
	\begin{equation}
	\label{eqn::fkupdate}
	\begin{aligned}
	&f_{\bm{w}(tK\tau+(r+1)\tau)}(x)-f^\star(x)\\
	&= U(f_{\bm{w}(tK\tau+r\tau)}(x)-f^\star)+o\left(\tau^2\right)\\
	\end{aligned}   
	\end{equation}
	
	We now derive an upper bound on the eigenvalue of $U$ when the stepsize is small enough. Since the range of each $\Pi_i$ has finite dimension in the function space, the operator norm of $\Pi_i$ is finite, and so is the operator norm of the multiplication of $\Pi_i$. We use $B_1$ to denote the maximum operator norm of $\sum_{j=1}^K\mathbf{Var}[\Pi_j]+\sum_{i> j}\left[\mathbb{E}\left[\Pi_i\right],\mathbb{E}\left[\Pi_j\right]\right]$ among all possible $\Pi$ operators. 
	
	For any function $f$ in the null space of $\Pi$, by assumption (v) in theorem 4.1, it is also in the null space of each $\Pi_i$, thus $\textbf{Var}\left[\Pi_k\right]f=\left[\mathbb{E}\left[\Pi_i\right],\mathbb{E}\left[\Pi_j\right]\right]f=0$ for any $i,j,k$. As a result, $Uf=f$.
	
	For any function in the orthogonal space of the null space of $\Pi_i$
	$$
	\left\|Uf\right\|\le \left(1+\frac{1}{2}\tau^2\eta^2B_1\right)e^{-\eta\tau K\lambda_{min}(\Pi)}\left\|f\right\|
	$$
	where $\lambda_{min}(\cdot)$ is the smallest nonzero eigenvalue of an operator. If $\eta$ and $\tau$ is small enough such that 
		\begin{equation}
		\label{eqn::sigmamaxusmallerthan1}
		\begin{aligned}
		\left(1+\frac{1}{2}\tau^2\eta^2B_1\right)e^{-\eta\tau K\lambda_{min}\left(\Pi\right)}< 1 
		\end{aligned}
		\end{equation}
	$f$ will decay to zero in the limit $\lim_{k\to\infty}U^kf=0$.

	As a simple estimate of \eqref{eqn::sigmamaxusmallerthan1}, when $\tau\eta$ is small enough, the leading term solution of equation \eqref{eqn::sigmamaxusmallerthan1} is $ \eta\tau\le\frac{2K\lambda_{min}\left(\Pi\right)}{K\lambda_{min}\left(\Pi\right)+B_1}$.
	
	
	By orthogonal decomposition, every function $f$ can be decomposed into $f=\Delta_0(f)+\Delta_{\perp}(f)$, where $\Delta_0$ is in the projection into the null space of $\Pi$ and $\Delta_{\perp}$ is in the projection into the orthogonal space of the null space of $\Pi$. By using relation \eqref{eqn::fkupdate} iteratively, $\Delta_{\perp}$ will shrink to zero, then we can obtain the limiting result:
	\begin{equation}
	\label{eqn::fkinfinity}
	\begin{aligned}
	f_{\bm{w}_{\infty}}(x)-f^\star(x)=\Delta_{0}(f_{\bm{w}_{0}}(x)-f^\star(x))+o\left(\eta^2\tau^2\right)\\
	\end{aligned}   
	\end{equation}
	We thus have:
	\begin{equation}
	\begin{aligned}
	f_{\bm{w}_{\infty}}(x)= f_{\bm{w}_{0}}(x)+ \Theta(x,X)\Theta^{-1}(X,X)(y^\star-f_{\bm{w}_{0}}(X)) + o\left(\tau^2\right)
	\end{aligned}
	\end{equation}
	where $f_{\bm{w}_{0}}(X)$ is a vector of dimension $n$ whose $i$-th component is $f_{\bm{w}_{0}}(x_i)$, and $\Theta(X,X)$ is an $n$ by $n$ matrix whose $ij$-th entry is $\left[\Theta(X,X)\right]_{ij}=\Theta\left(x_i,x_j\right)$. This result shows that although each mode is trained only on a part of clients, when we use block coordinate descent rule, the limiting behavior of individual mode is similar to the case where we train each mode on the entire dataset, with a small higher order discrepancy term that depends on local epochs and data heterogeneity.
	
	As a result, each mode can be regarded as independent draw from Gaussian $\mathcal{GP}\left(m(x),k(x,x^\prime)\right)$
	\begin{equation*}
	m(x)=\Theta(x,X)\Theta^{-1}(X,X)y^\star
	\end{equation*}
	and 
	\begin{equation*}
	\begin{aligned}
	&k\left(x,x^\prime\right)= \mathcal{K}\left(x,x^\prime\right)+\Theta(x,X)\Theta^{-1}\mathcal{K}\Theta^{-1}\Theta(X,x^\prime)\\
	&-\left(\Theta(x,X)\Theta^{-1}\mathcal{K}\left(X,x^\prime\right)+\Theta(x^\prime,X)\Theta^{-1}\mathcal{K}\left(X,x\right)\right)
	\end{aligned}
	\end{equation*}
	where $\mathcal{K}(\cdot,\star)=\mathbb{E}_{\bm{w}(0)\sim p_{init}}\left[f_{\bm{w}_0}(\cdot)f_{\bm{w}_0}(\star)\right]$ is the kernel at initialization. As a special case, if $\mathcal{K}(\cdot,\star)=\sigma^2\Theta(\cdot,\star)$, which is the case of toy example in section 5.1  of the main text, the posterior variance is given by:
	\begin{equation*}
	k(x,x^\prime)=\sigma^2\left(\Theta\left(x,x^\prime\right)-\Theta(x,X)\Theta^{-1}(X,X)\Theta(X,x^\prime)\right).
	\end{equation*}

	This completes our discussion on training linearized model. 
	
	\subsection{Convergence and change of neural tangent kernel}
	\label{ap::trainneuralnet}
	Training infinite width neural network by small-stepsize gradient descent is equivalent to training a linear model with fixed kernel. In this subsection, we prove similarly that the tangent kernel remains fixed during \name, following the derivations of \cite{ntkgaussian}. If the neural tangent kernel is fixed, we can adopt the analysis of sec \ref{ap::trainlinearmodel} to neural network by taking $\Theta$ to be the limiting kernel. From the result in sec \ref{ap::trainlinearmodel}, we can prove theorem \ref{thm::ntkgaussian} in the main text.
	
	For NTK initialization \cite{ntk} $f_{\bm{w}_0}$ converges to a Gaussian process whose mean is zero and whose kernel is denoted as $\mathcal{K}$. Then for any $\delta_0$, there exists $R_0$, and $n_0$ such that for every $l\ge l_0$, $\left\|g(\bm{w}_0)\right\|_2\le R_0$ with probability at least $1-\delta_0$.
	Also, in NTK setting, it is proved as Lemma 1 in \cite{ntkspectrum} that the operator norm of Hessian of function $\left\|\frac{1}{\sqrt{l}}\nabla^2 g(\bm{w})\right\|_{op}$ is bounded above by $C_1$, and that $C_1\to 0$ in the limit $l\to\infty$. 
	\begin{theorem}
		(Convergence of global training error) Under the assumptions listed below:
		\begin{enumerate}
		    \item $\eta_0\tau$ is small enough such that $\kappa_2$ defined in \eqref{eqn::defofkappa2} is smaller than $1$.
		    \item The width of neural network $l$ is large enough.
		\end{enumerate}
		for positive constants $R_0,\ \delta_0>0$, the following holds with probability $1-\delta_0$:
		
		\begin{equation}
		\label{eqn::convergence}
		\left\{\begin{aligned}
		&\left\|g(\bm{w}(tK\tau))\right\|_2\le e^{-\frac{\eta_0\lambda_{m}}{3\bar{n}}tK\tau}R_0+o(\eta_0^2\tau^2)\\
		&\left\|\bm{w}(tK\tau)-\bm{w}(0)\right\|_2\le \frac{3\eta_0\bar{n} R_0l^{-\frac{1}{2}}\left(1+\kappa_2\right)C_2}{\lambda_m}\left(1- e^{-\frac{\eta_0\lambda_m}{3\bar{n}}t \tau}\right)+o(\eta_0^2\tau^2)
		\end{aligned}\right.
		\end{equation}
		$\forall t=1,2,...$. 
		And also, $\Theta(\bm{w}(tK\tau))$ is not very far away from $\Theta$:
		$$
		\left\|\Theta(\bm{w}(tK\tau))-\Theta\right\|\le O\left(l^{-\frac{1}{2}}\right)
		$$
		
		The notations is defined as following: $\bar{n}$ is the number of datapoints in one communication round, $C_2$ is a constant defined as the summation of all $C^{(l)}_i$'s in lemma \ref{lm::jacobianlip},  $C_2=\sum_{i=1}^NC^{(l)}_i$,
		and $\kappa_2$ is an $O(\eta_0\tau)$ constant defined in \eqref{eqn::defofkappa2}:
		\begin{equation}
		\label{eqn::defofdelta2}
		 \begin{aligned}
		&\kappa_2=-1+\exp\left(\eta_0\tau C_2+\eta_0^2\tau^2\left(2C_2^4+8\frac{R_0}{\sqrt{l}}C_1C_2^2\right)\right)\\
		&=O(\tau\eta_0)
		\end{aligned}   
		\end{equation}
		
	\end{theorem}
	Note that the second equation in \eqref{eqn::convergence} is the result of theorem \ref{thm::traininglossconverge}.
	
	\begin{proof}
	We will prove \eqref{eqn::convergence} by induction. At $t=0$, the second equation is trivially true, and the first reduces to $\left\|g(\bm{w}(0))\right\|_2\le R_0$, which is also true with probability at least $1-\frac{\delta_0}{10}$. We choose proper constants so that Lemma \ref{lm::jacobianlip} holds with probability at least $1-\frac{\delta}{10}$ over random initialization. 
		Now assume \eqref{eqn::convergence} hold for $t=1,2,...\xi$, we will prove they still hold for $t=\xi+1$. For clarity, we call this induction the induction across ages.
		
		\paragraph{Communication rounds within one age}
		We will prove the following by another level of induction:
		\begin{equation}
		\label{eqn::convergencewithinage}
		\left\{\begin{aligned}
		&\left\|g(\bm{w}(\xi K\tau+r\tau))\right\|_2\le \left(1+\kappa_2\right)\left\|g(\bm{w}(\xi K\tau))\right\|_2+o(\tau^2)\\
		&\left\|\bm{w}(\xi K\tau+r\tau)-\bm{w}(0)\right\|_2\le\frac{3\eta_0 R_0l^{-\frac{1}{2}}\left(1+\kappa_2\right)C_2}{\lambda_m}\left(1- e^{-\frac{\eta_0\lambda_m}{3\bar{n}}\xi \tau}\right)+o(\tau^2)
		\end{aligned}\right.		\quad \forall r=1,2,...K
		\end{equation}
		We call this induction the induction across communication rounds.
		
		Obviously, \eqref{eqn::convergencewithinage} holds for $r=0$, which is the inductive assumption of induction across different ages. Now we assume that \eqref{eqn::convergencewithinage} hold for communication rounds $r^\prime=1,2,...r$, and prove the inequalities for $r^\prime=r+1$.
		
		As discussed before, we denote $\bm{w}(\xi K\tau+r\tau)$ the weights after the server update step of the $r$-th communication round of age $\xi$. In the beginning of next communication round, $\bm{w}(\xi K\tau+r\tau)$ is sent to all clients for some selected strata. Under gradient descent rule, the client update can be approximated by the differential equation:
		\begin{equation}
		\begin{aligned}
		&\frac{d}{d{t_i}}\bm{w}^{(i)}(\xi K\tau+r\tau+t_i)\\
		&=-\frac{\eta}{n_i} J_i(\bm{w}^{(i)}(\xi K\tau+r\tau+t_i))g_{i}(\bm{w}^{(i)}(\xi K\tau+r\tau+t_i))
		\end{aligned}
		\end{equation}
		where $t_i$ is the local training epoch of client $i$ at communication round $r$, $\bm{w}^{(i)}(\xi K\tau+r\tau+t_i)$ is the weight on client $i$ starting from $\bm{w}^{(i)}(\xi K\tau+r\tau)=\bm{w}(\xi K\tau+r\tau)$, and other notations are defined accordingly. As a result, the dynamics of $g$ is given by:
		\begin{equation}
		\begin{aligned}
		&\frac{d}{d{t_i}}g(\bm{w}^{(i)}(\xi K\tau+r\tau+t_i))\\
		&=-\frac{\eta}{n_i} J(\bm{w}^{(i)}(\xi K\tau+r\tau+t_i))^TJ_i(\bm{w}^{(i)}(\xi K\tau+r\tau+t_i))g_{i}(\bm{w}^{(i)}(\xi K\tau+r\tau+t_i))
		\end{aligned}
		\end{equation}
		We denote $\Theta_i$ as $\Theta_i(t_i)=\frac{1}{l}J(\bm{w}^{(i)}(\xi K\tau+r\tau+t_i))^TJ_i(\bm{w}^{(i)}(\xi K\tau+r\tau+t_i))\mathcal{P}_i$, then the solution to the differential equation is:
		$$
		g(\bm{w}^{(i)}(\xi K\tau+r\tau+t_i))=\text{exp}\left(-\frac{\eta_0}{n_i}\int_{0}^{\tau}\Theta_i(t_i)dt_i\right)g(\bm{w}(\xi K\tau+r\tau))
		$$
		By integral mean value theorem, there exists a $\bar{t_i}\in[0,\tau]$ such that $\int_{0}^{\tau}\Theta_i(t_i)dt_i=\tau\Theta_i(\bar{t_i})$, thus 
		$$
		g(\bm{w}^{(i)}(\xi K\tau+r\tau+\tau))=\text{exp}\left(-\frac{\eta_0}{n_i}\Theta_i(\bar{t_i})\tau\right)g(\bm{w}(\xi K\tau+r\tau))
		$$
		After $\tau$ local epochs, all clients will send model to server, and the server performs $server\_update$ to calculate $\bm{w}(\xi K\tau+(r+1)\tau)$ based on received local modes. As defined before, $S(\xi,r)$ is the indices of clients that download model $\bm{w}(\xi K\tau+r\tau)$ in communication round $r$ of age $\xi$, then the server update function can be written as:
		$$
		\bm{w}(\xi K\tau+(r+1)\tau)=\frac{1}{\sum_{i\in S(\xi,r)}n_i}\sum_{i\in S(\xi,r)}n_i\bm{w}^{(i)}(\xi K\tau+r\tau+\tau)
		$$ 
		The averaged error function is given by:
		\begin{equation}
		\label{eqn::paraavg2funcavg}
		\begin{aligned}
		&\frac{1}{\sum_{i\in S(\xi,r)}n_i}\sum_{i\in S(\xi,r)}n_ig(\bm{w}^{(i)}(\xi K\tau+r\tau+\tau))\\
		&=g(\bm{w}(\xi K\tau+(r+1)\tau))+\eta_0^2\tau^2v_1^{(\xi,r)}+o(\eta_0^2\tau^2)
		\end{aligned}		
		\end{equation}
		
		We then try to estimate the averaged error vector by the function parametrized by averaged weights and bound their difference. To do so, we use a simple notation $v^{(\xi,r)}_1$ to denote all the second order terms in \eqref{eqn::paraavg2funcavg}:
		$$
		\begin{aligned}
		v^{(\xi,r)}_1 = \frac{1}{\eta_0^2\tau^2\sum_{i\in S(\xi,r)}n_i}\sum_{i\in S(\xi,r)}n_i\Delta \bm{w}_i^T \nabla^2g(\bm{w}(\xi K\tau+(r+1)\tau))\Delta \bm{w}_i
		\end{aligned}
		$$
		where 
		$$
		\begin{aligned}
		&\Delta\bm{w}_i=\bm{w}^{(i)}(\xi K\tau+r\tau+\tau)-\bm{w}(\xi K\tau+(r+1)\tau)\\
		&=\bm{w}^{(i)}(\xi K\tau+r\tau+\tau)-\bm{w}(\xi K\tau+r\tau)-\left(\bm{w}(\xi K\tau+(r+1)\tau)-\bm{w}(\xi K\tau+r\tau)\right)
		\end{aligned}
		$$
		For each client $i$, by integral mean value theorem, there exists a $\tilde{t_i}\in[0,\tau]$, such that:
		$$
		\begin{aligned}
		&\bm{w}^{(i)}(\xi K\tau+r\tau+\tau)-\bm{w}(\xi K\tau+r\tau)=-\frac{\eta\tau}{n_i} J_i(\bm{w}^{(i)}(\xi K\tau+r\tau+\tilde{t_i}))g_{i}(\bm{w}^{(i)}(\xi K\tau+r\tau+\tilde{t_i}))
		\end{aligned}
		$$
		As a result:
		$$
		\begin{aligned}
		&\Delta \bm{w}_i=-\frac{\eta\tau}{n_i} J_i(\bm{w}^{(i)}(\xi K\tau+r\tau+\tilde{t_i}))g_{i}(\bm{w}^{(i)}(\xi K\tau+r\tau+\tilde{t_i}))\\
		&+\frac{\eta\tau}{\sum_{i\in S(\xi,r)}n_i}\left(\sum_{j\in S(\xi,r)}J_j(\bm{w}^{(j)}(\xi K\tau+r\tau+\tilde{t_i}))g_{j}(\bm{w}^{(j)}(\xi K\tau+r\tau+\tilde{t_i}))\right)\\
		&=-\frac{\eta\tau}{n_i} J_i(\bm{w}(\xi K\tau+r\tau))g_{i}(\bm{w}(\xi K\tau+r\tau))\\
		&+\frac{\eta\tau}{\sum_{i\in S(\xi,r)}n_i}\left(\sum_{j\in S(\xi,r)}J_j(\bm{w}(\xi K\tau+r\tau))g_{j}(\bm{w}(\xi K\tau+r\tau))\right)+o(\tau)\\
		&=\frac{\eta_0\tau}{\sqrt{l}} D_{1}^{(\xi,r,i)}g(\bm{w}(\xi K\tau+r\tau))+o(\tau)
		\end{aligned}
		$$
		where $D_{1}^{(\xi,r,i)}$ is an operator defined as:
		$$
		\begin{aligned}
		D_{1}^{(\xi,r,i)}=-\frac{1}{n_i\sqrt{l}} J_i(\bm{w}(\xi K\tau+r\tau))\mathcal{P}_i+\frac{1}{\sum_{i\in S(\xi,r)}n_i\sqrt{l}}\left(\sum_{j\in S(\xi,r)}J_j(\bm{w}(\xi K\tau+r\tau))\mathcal{P}_j\right)
		\end{aligned}
		$$
		And its operator norm is upper bounded by $2C_2$. We replace $\tilde{t}_i$ in the third equality by $0$. Due to continuity, the difference is proportional to $\tilde{t}_i\le \tau$, thus can be absorbed into $o(\tau)$ term after multiplying $\tau$. Therefore, $v_1$ is 
		$$
		\begin{aligned}
		v_1^{(\xi,r)}=\frac{1}{\sum_{i\in S(\xi,r)}n_il}\sum_{i\in S(\xi,r)}n_i\left(g(\bm{w}(\xi K\tau+r\tau))\right)^T\left(D_1^{(\xi,r,i)}\right)^T\nabla^2g D_1^{(\xi,r,i)}g(\bm{w}(\xi K\tau+r\tau))
		\end{aligned}
		$$
		We can define a new operator $D_1^{(\xi,r)}$ as:
		$$
		D_1^{(\xi,r)}=\frac{\left(g(\bm{w}(\xi K\tau+r\tau))\right)^T}{\sum_{i\in S(\xi,r)}n_i}\sum_{i\in S(\xi,r)}n_i\left(D_1^{(\xi,r,i)}\right)^T\nabla^2g D_1^{(\xi,r,i)}
		$$
		The operator norm of $D_1^{(\xi,r)}$ is thus bounded by:
		$$
		\begin{aligned}
		&\left\|D_1^{(\xi,r)}\right\|_{op}\le \left\|g(\bm{w}(\xi K\tau+r\tau))\right\|_2\frac{1}{\sum_{i\in S(\xi,r)}n_i}\sum_{i\in S(\xi,r)}n_i\left\|\left(D_1^{(\xi,r,i)}\right)^T\nabla^2g D_1^{(\xi,r,i)}\right\|_{op}
		\end{aligned}
		$$
		By inductive assumption \eqref{eqn::convergencewithinage}, both $\left\|g(\bm{w}(\xi K\tau+r\tau))\right\|_2$ and $\left\|\frac{1}{\sqrt{l}}\nabla^2g\right\|$ are upper bounded, also by lemma \ref{lm::jacobianlip}, $\left\|D_1^{(\xi,r,i)}\right\|_{op}$ is upper bounded by $2C_2$, thus $\|D_1^{(\xi,r)}\|_{op}$ is upper bounded by:
		$$
		\|D_1^{(\xi,r)}\|_{op}\le \frac{2}{\sqrt{l}}R_0C_1\left(2C_2\right)^2
		$$
		
		As a result, error vector $g(\bm{w}(\xi K\tau+(r+1)\tau))$ can be calculated as: 
		\begin{equation}
		\label{eqn::gupdateinoneage}
		\begin{aligned}
		&g(\bm{w}(\xi K\tau+(r+1)\tau))\\
		&=\frac{1}{\sum_{i\in S(\xi,r)}n_i}\sum_{i\in S(\xi,r)}n_i\text{exp}\left(-\frac{\eta_0}{n_i}\Theta_i(\bar{t_i})\tau\right)g(\bm{w}(\xi K\tau+r\tau))\\
		&-\eta^2\tau^2D_1^{(\xi,r)}g(\bm{w}(\xi K\tau+r\tau))+o(\tau^2)\\
		&=\text{exp}\left(-\eta_0\tau\frac{1}{\sum_{i\in S(\xi,r)}n_i}\sum_{i\in S(\xi,r)}\Theta_i(\bar{t_i})\right)g(\bm{w}(\xi K\tau+r\tau))\\
		&+\eta_0^2\tau^2D_2^{(\xi,r)}g(\bm{w}(\xi K\tau+r\tau))+o(\tau^2)\\
		&=\left(1+\eta_0^2\tau^2D_2^{(\xi,r)}\right)\text{exp}\left(-\eta_0\tau\frac{1}{\sum_{i\in S(\xi,r)}n_i}\sum_{i\in S(\xi,r)}\Theta_i(\bar{t_i})\right)g(\bm{w}(\xi K\tau+r\tau))+o(\tau^2)
		\end{aligned}
		\end{equation}
		
		where $D_2^{(\xi,r)}$ is defined as
		\begin{equation}
		\label{eqn::d2def}
		\begin{aligned}
		&D_2^{(\xi,r)}=\frac{1}{\sum_{i\in S(\xi,r)}n_i}\sum_{i\in S(\xi,r)}n_i\Theta_i^2(\bar{t_i})-\left(\frac{1}{\sum_{i\in S(\xi,r)}n_i}\sum_{i\in S(\xi,r)}n_i\Theta_i(\bar{t_i})\right)^2-D_1^{(\xi,r)}
		\end{aligned}
		\end{equation}

		The second equality of \eqref{eqn::gupdateinoneage} comes from lemma \ref{lm::varofoperator}, and the third combines $\eta_0^2\tau^2D_2^{(\xi,r)}$ into product term.
		
		We can replace $\bar{t_i}$ in \eqref{eqn::d2def} by 0, since the difference can be absorbed into $o(\tau^2)$ terms. By induction hypothesis, $\left\|\Theta_i(0)\right\|_{op}$ is bounded by		$\left(C_i^{(l)}\right)^2\le C_2^2$. Hence, the operator norm of		$D_2^{(\xi,r)}$ is also bounded by constants		$\kappa_2^{(\xi,r)}=2C_2^4+8\frac{R_0}{\sqrt{l}}C_1C_2^2$. Therefore:
		$$
		\begin{aligned}
		& \left\|g(\bm{w}(\xi K\tau+(r+1)\tau))\right\|_2\\
		&\le (1+\eta_0^2\tau^2\kappa_2^{(\xi,r)})\left\|\text{exp}\left(-\eta_0\tau\frac{1}{\sum_{i\in S(\xi,r)}n_i}\sum_{i\in S(\xi,r)}\Theta_i(\bar{t_i})\right)\right\|_{op}\\
		&\left\|g(\bm{w}(\xi K\tau+r\tau))\right\|_2+o(\tau^2)\\
		&\le \exp\left(\eta_0^2\tau^2\kappa_2^{(\xi,r)}\right)\text{exp}\left(\eta_0\tau\sum_{i\in S(\xi,r)}C^{(l)}_i\right)\left\|g(\bm{w}(\xi K\tau+r\tau))\right\|_2+o(\tau^2)\\
		&\le \cdots\\
		&\le \exp\left(\eta_0\tau\sum_{r^\prime}^r\sum_{i\in S(\xi,r)}C_i^{(l)}+\eta_0^2\tau^2 \kappa_2^{(\xi,r^\prime)}\right)\left\|g(\bm{w}(\xi K\tau))\right\|_2+o(\tau^2)\\
		&\le (1+\kappa_2)\left\|g(\bm{w}(\xi K\tau))\right\|_2+o(\tau^2)\\
		\end{aligned}
		$$
		where 
		\begin{equation}
		\label{eqn::defofkappa2}
		\begin{aligned}
		&\kappa_2=-1+\exp\left(\eta_0\tau C_2+\eta_0^2\tau^2\left(2C_2^4+8\frac{R_0}{l}C_1C_2^2\right)\right)
		\end{aligned}
		\end{equation}
		
		We thus prove the first equation in \eqref{eqn::convergencewithinage}. 
		
		For the second one, we should consider the dynamics of $\bm{w}$. On each client's side, 
		$$
		\frac{d}{d{t_i}}\bm{w}^{(i)}(\xi K\tau+r\tau+t_i)=-\frac{\eta}{n_i} J_i(\bm{w}^{(i)}(\xi K\tau+r\tau+t_i))g_{i}(\bm{w}^{(i)}(\xi K\tau+r\tau+t_i))
		$$
		as a result:
		$$
		\begin{aligned}
		&\frac{d}{d{t_i}}\left\|\bm{w}^{(i)}(\xi K\tau+r\tau+t_i)-\bm{w}_0\right\|_2\\
		&\le \left\|\frac{d}{d{t_i}}\bm{w}^{(i)}(\xi K\tau+r\tau+t_i)\right\|_2\\
		&=\frac{\eta_0}{ln_i}\left\| J_i(\bm{w}^{(i)}(\xi K\tau+r\tau+t_i))\mathcal{P}_ig(\bm{w}^{(i)}(\xi K\tau+r\tau+t_i))\right\|_2\\
		&\le \frac{\eta_0}{ln_i}\left\| J_i(\bm{w}^{(i)}(\xi K\tau+r\tau+t_i))\mathcal{P}_i\right\|_{op}\left\|g(\bm{w}^{(i)}(\xi K\tau+r\tau+t_i))\right\|_2\\
		\end{aligned}
		$$
		We use another layer of induction here. If $\left\|\bm{w}^{(i)}(\xi K\tau+r\tau)(t_i^\prime)-\bm{w}_0\right\|_2\le \frac{3 R_0l^{-\frac{1}{2}}\left(1+\kappa_2\right)C_2}{\lambda_m}\left(1- e^{-\frac{\eta_0\lambda_m}{3}\xi K\tau}\right)+\eta_0R_0C^{(l)}_it_i^\prime l^{-\frac{1}{2}}n_i^{-1}\left(1+\kappa_2\right)\text{exp}\left(-\frac{1}{3}\eta_0\lambda_{m}\xi K\tau\right)$, for all $t_i^\prime<t_i$, then it's also true that \\ $\left\|\bm{w}^{(i)}(\xi K\tau+r\tau+t_i)-\bm{w}_0\right\|_2\le 3 R_0l^{-\frac{1}{2}}\left(1+\kappa_2\right)C_2\lambda_m^{-1}$, by lemma \ref{lm::jacobianlip}, we know:
		$$
		l^{-\frac{1}{2}}\left\| J_i(\bm{w}^{(i)}(\xi K\tau+r\tau+t_i^\prime))\mathcal{P}_i\right\|_{op}\le C^{(l)}_i
		$$
		for any $t_i^\prime<t_i$. By integral mean value theorem, we finally have:
		$$
		\begin{aligned}
		&\left\|\bm{w}^{(i)}(\xi K\tau+r\tau+t_i)-\bm{w}(0)\right\|_2-\left\|\bm{w}^{(i)}(\xi K\tau+r\tau)-\bm{w}(0)\right\|_2\\
		&\le \eta_0R_0C^{(l)}_it_i l^{-\frac{1}{2}}n_i^{-1}\left(1+\kappa_2\right)\text{exp}\left(-\frac{1}{3\bar{n}}\eta_0\lambda_{m}\xi \tau\right)
		\end{aligned}
		$$
		The upper bound thus also holds for $t_i$, $\forall t_i\in[0,\tau]$. We can set $t_i=\tau$:
		$$
		\begin{aligned}
		&\left\|\bm{w}^{(i)}(\xi K\tau+r\tau+\tau)-\bm{w}(0)\right\|_2-\left\|\bm{w}^{(i)}(\xi K\tau+r\tau)-\bm{w}(0)\right\|_2\\
		&\le \eta_0R_0C^{(l)}_i\tau l^{-\frac{1}{2}}n_i^{-1}\left(1+\kappa_2\right)\text{exp}\left(-\frac{1}{3\bar{n}}\eta_0\lambda_{m}\xi \tau\right)
		\end{aligned}
		$$
		By convexity of the $\ell_2$ norm, we know that
		$$
		\begin{aligned}
		&\left\|\bm{w}(\xi K\tau+(r+1)\tau)-\bm{w}(0)\right\|_2-\left\|\bm{w}(\xi K\tau+r\tau)-\bm{w}(0)\right\|_2\\
		&\le \eta_0\tau R_0l^{-\frac{1}{2}}\left(1+\kappa_2\right)\frac{\sum_{i\in S(\xi,r)}C^{(l)}_i}{\sum_{i\in S(\xi,r)}n_i}\text{exp}\left(-\frac{1}{3\bar{n}}\eta_0\lambda_{m}\xi \tau\right)\\
		&\le \eta_0\tau R_0l^{-\frac{1}{2}}\left(1+\kappa_2\right)C_2\text{exp}\left(-\frac{1}{3\bar{n}}\eta_0\lambda_{m}\xi \tau\right)
		\end{aligned}
		$$
		where $C_2=\sum_{i=1}^NC^{(l)}_i$.
		
		As a result:
		$$
		\begin{aligned}
		&\left\|\bm{w}(\xi K\tau+(r+1)\tau)-\bm{w}(0)\right\|_2 \\
		&\le r\eta_0\tau R_0l^{-\frac{1}{2}}\left(1+\kappa_2\right)C_2\text{exp}\left(-\frac{1}{3\bar{n}}\eta_0\lambda_{m}\xi \tau\right)+\left\|\bm{w}(\xi K\tau)-\bm{w}(0)\right\|_2\\
		&\le \frac{3\eta_0 R_0l^{-\frac{1}{2}}\left(1+\kappa_2\right)C_2}{\lambda_m}\left(1- e^{-\frac{\eta_0\lambda_m}{3\bar{n}}(\xi+1) \tau}\right)
		\end{aligned}
		$$
		Thus the second equation in \eqref{eqn::convergencewithinage} holds. Also, the second equation in \eqref{eqn::convergence} holds.
		
		\paragraph{Exponential decrease of error}
		Now we come to the induction across different ages, in which we will prove that the norm of error vector decreases at an exponential rate. From \eqref{eqn::gupdateinoneage}:
		$$
		\begin{aligned}
		&g(\bm{w}((\xi+1)K\tau))\\
		&=\left(\prod_{r=1}^K\left(1+\eta_0^2\tau^2D_2^{(\xi,r)}\right)\text{exp}\left(-\eta_0\tau\frac{1}{\sum_{i\in S(\xi,r)}n_i}\sum_{i\in S(\xi,r)}\Theta_i(\bar{t_i})\right)\right)g(\bm{w}(\xi K\tau))+o(\tau^2)\\
		&=\left(1+\eta_0^2\tau^2D_3^{(\xi)}\right)\text{exp}\left(-\eta_0\tau\sum_r\frac{1}{\sum_{i\in S(\xi,r)}n_i}\sum_{i\in S(\xi,r)}\Theta_i(\bar{t_i})\right)g(\bm{w}(\xi K\tau))+o(\tau^2)\\
		\end{aligned}
		$$
		where $D_3^{(\xi)}$ is defined as:
		$$
		\begin{aligned}
		&D_3^{(\xi)}=\sum_{r=1}^KD_2^{(\xi,r)}+\frac{1}{2}\sum_{\beta>\gamma}\left[\frac{1}{\sum_{i\in S(\xi,r_{\beta})}n_i}\sum_{i\in S(\xi,r_{\beta})}\Theta_i(\bar{t_i}),\frac{1}{\sum_{i\in S(\xi,r_{\gamma})}n_i}\sum_{i\in S(\xi,r_{\gamma})}\Theta_i(\bar{t_i}) \right]
		\end{aligned}
		$$
		We have shown that $\left\|D_2\right\|$ is upper bounded, also by lemma \ref{lm::jacobianlip} and second equation in \eqref{eqn::convergence}, $\left\|D_3^{(\xi)}\right\|_{op}$ is upper bounded by constant $\kappa_3^{(\xi)}$: 
		$\left\|D_3^{(\xi)}\right\|_{op}\le \kappa_3^{(\xi)}$.
		
		$\kappa_3=\max_{\xi}\kappa_3^{(\xi)}$ is a $O(\tau^2)$ constant defined as:
		$$
		\begin{aligned}
		&\kappa_3=\eta_0^2\tau^2\left(2C_2^4+8\frac{R_0}{\sqrt{l}}C_1C_2^2 \right)+\eta_0^2\tau^2\frac{N(N-1)}{2}C_2^4\\
		&=O(\tau^2\eta_0^2)
		\end{aligned}
		$$
		
		By assumption $\sum_{i\in S(\xi,r)}n_i=\bar{n}$ is a constant, thus we can rewrite $g(\bm{w}((\xi+1)K\tau))$ as:
		\begin{equation}
		\begin{aligned}
		&g(\bm{w}((\xi+1)K\tau))\\
		&=\left(1+\eta_0^2\tau^2D_3^{(\xi)}\right)\text{exp}\left(-\frac{\eta_0\tau}{\bar{n}}\sum_r\sum_{i\in S(\xi,r)}\Theta_i(\bar{t_i})\right)g(\bm{w}(\xi K\tau))+o(\tau^2)\\
		&=\left(1+\eta_0^2\tau^2D_3^{(\xi)}\right)\text{exp}\left(-\frac{\eta_0\tau}{\bar{n}}\sum_{i=1}^N\Theta_i(\bar{t_i})\right)g(\bm{w}(\xi K\tau))+o(\tau^2)\\
		\end{aligned}
		\end{equation}
		
		We will now prove that, with high probability, 
		$$
		\lambda_{min}\left(\sum_{i=1}^{N}\Theta_i(\bar{t_i}) \right)\ge\frac{\lambda_m}{2}
		$$
		The smallest eigenvalue of summation of kernel is:
		$$
		\begin{aligned}
		&\lambda_{min}\left(\sum_{i=1}^{N}\Theta_i(\bar{t_i}) \right)\\
		&=\lambda_{min}\left(\Theta+\frac{1}{l}J(\bm{w}(0))J(\bm{w}(0))^T-\Theta+\sum_{i=1}^{N}\Theta_i(\bar{t_i})-\frac{1}{l}J(\bm{w}(0))J(\bm{w}(0))^T \right)\\
		&\ge\lambda_{min}\left(\Theta \right)-\left\|\frac{1}{l}J(\bm{w}(0))J(\bm{w}(0))^T-\Theta\right\|_{op}-\left\|\sum_{i=1}^{N}\Theta_i(\bar{t_i})-\frac{1}{l}J(\bm{w}(0))J(\bm{w}(0))^T\right\|_{op}\\
		\end{aligned}
		$$
		The first term is just $\lambda_m$, the second term $\left\|J(\bm{w}(0))J(\bm{w}(0))^T-\Theta\right\|_{op}$ can be smaller than $\frac{\lambda_m}{3}$ by \cite{ntk} when the width $l$ is large enough with probability at least $1-\frac{\delta_0}{5}$. Now we will bound the third term:
		$$
		\begin{aligned}
		&\left\|\sum_{i=1}^{N}\Theta_i(\bar{t_i})-\frac{1}{l}J(\bm{w}(0))J(\bm{w}(0))^T\right\|_{op}\\
		&=\frac{1}{l}\left\|\sum_{i=1}^{N}J(\bm{w}(\bar{\bar{t_i}}))J_i(\bm{w}(\bar{\bar{t_i}}))^T\mathcal{P}_i-J(\bm{w}(0))\sum_{i=1}^{N}J_i(\bm{w}(0))^T\mathcal{P}_i\right\|_{op}\\
		&=\frac{1}{l}\left\|\sum_{i=1}^{N}\left(J(\bm{w}(\bar{\bar{t_i}}))J_i(\bm{w}(\bar{\bar{t_i}}))^T\mathcal{P}_i-J(\bm{w}(0))J_i(\bm{w}(0))^T\mathcal{P}_i\right)\right\|_{op}\\
		&\le\frac{1}{l}\sum_{i=1}^{N}\left\|J(\bm{w}(\bar{\bar{t_i}}))J_i(\bm{w}(\bar{\bar{t_i}}))^T\mathcal{P}_i-J(\bm{w}(0))J_i(\bm{w}(0))^T\mathcal{P}_i\right\|_{op}\\
		&\le\frac{1}{l}\sum_{i=1}^{N}\left\|J(\bm{w}(\bar{\bar{t_i}}))J_i(\bm{w}(\bar{\bar{t_i}}))^T\mathcal{P}_i-J(\bm{w}(\bar{\bar{t_i}}))J_i(\bm{w}(0))^T\mathcal{P}_i\right\|_{op}\\
		&+\left\|J(\bm{w}(\bar{\bar{t_i}}))J_i(\bm{w}(0))^T\mathcal{P}_i-J(\bm{w}(0))J_i(\bm{w}(0))^T\mathcal{P}_i\right\|_{op}\\
		&\le\frac{1}{l}\sum_{i=1}^{N}\left\|J(\bm{w}(\bar{\bar{t_i}}))\right\|_{F}C^{(l)}_i\left\|\bm{w}(\bar{\bar{t_i}})-\bm{w}_{0}\right\|_2+\left\|J_i(\bm{w}(0))^T\mathcal{P}_i\right\|_{F}C^{(l)}\left\|\bm{w}(\bar{\bar{t_i}})-\bm{w}_{0}\right\|_2\\
		&\le \frac{\lambda_m}{6}
		\end{aligned}
		$$
		when the width $l$ is large enough. We use $\bar{\bar{t_i}}$ to denote $\bar{t_i}+\xi K\tau+r\tau$
		
		Thus we have:
		$$
		\left\|g(\bm{w}((n+1)K\tau))\right\|_2\le \left(1+\kappa_3\right)\text{exp}\left(-\frac{\eta_0\tau}{\bar{n}}\frac{\lambda_m}{2}\right)\left\|g(\bm{w}(\xi K\tau))\right\|_2
		$$
		with probability at least $1-\frac{\delta_0}{2}$. Since $\kappa_3$ is $O(\eta^2\tau^2)$, when $\eta_0\tau$ is small enough we have:
		$$
		\left\|g(\bm{w}((n+1)K\tau))\right\|_2\le \text{exp}\left(-\frac{\eta_0\tau}{\bar{n}}\frac{\lambda_m}{3}\right)\left\|g(\bm{w}(\xi K\tau))\right\|_2
		$$
		By union bound of probabilities, we thus complete the proof of the first equation in \eqref{eqn::convergence}. 
		
		The upper bound of change of neural tangent kernel during training follows from directly Lemma \ref{lm::jacobianlip} and second equation of \eqref{eqn::convergence}.
		
	\end{proof}
\subsection{Difference of neural network training and linear model training}
\label{sec:difbound}
In this section we will bound the difference between the training of neural network and that of linear model, and show that the bound goes to zero in the limit of infinitely wide network. Then combining with derivations in Sec. \ref{ap::trainlinearmodel}, we can prove theorem \ref{thm::ntkgaussian}. 

We use $g^{lin}$ to denote the error vector when training linear model with limiting kernel $\Theta$. 

We start from the dynamics on individual client. We denote $\Theta_{i0}$ as $\Theta_{i0}=\frac{1}{l}J(\bm{w}(0))^TJ_i(\bm{w}(0))\mathcal{P}_i$. Similar to \cite{ntkgaussian}, on client $i$, we have:
\[
\begin{aligned}
&\frac{d}{dt}\left(\exp{\left(\eta_0\frac{\Theta_{i0}}{n_i}t\right)}\left(g^{lin}(\bm{w}^{(i)}(\xi K\tau+r\tau+t))-g(\bm{w}^{(i)}(\xi K\tau+r\tau+t))\right)\right)\\
&=\eta_0\exp{\left(\eta_0\frac{\Theta_{i0}}{n_i}t\right)}\left(\Theta_i(\bm{w}^{(i)}(\xi K\tau+r\tau+t))-\Theta_{i0})g(t)\right)
\end{aligned}
\]
Integrating both sides from $0$ to $\tau$, we have:
\[
\begin{aligned}
&g^{lin}(\bm{w}^{(i)}(\xi K\tau+r\tau+\tau))-g(\bm{w}^{(i)}(\xi K\tau+r\tau+\tau))\\
&=\exp{\left(-\eta_0\frac{\Theta_{i0}}{n_i}\tau\right)}\left(g^{lin}(\bm{w}(\xi K\tau+r\tau))-g(\bm{w}(\xi K\tau+r\tau))\right)\\
&+\exp{\left(-\eta_0\frac{\Theta_{i0}}{n_i}\tau\right)}\eta_0\int_0^{\tau}\exp{\left(\eta_0\frac{\Theta_{i0}}{n_i}s\right)}\left(\Theta_i(\bm{w}^{(i)}(\xi K\tau+r\tau+s))-\Theta_{i0})g(s)\right)ds
\end{aligned}
\]

Then we take weighted average on both sides. Since
\[
\bm{w}(\xi K\tau+(r+1)\tau)=\frac{1}{\sum_{i\in S(\xi,r)}n_i}\sum_{i\in S(\xi,r)}n_i\bm{w}^{(i)}(\xi K\tau+r\tau+\tau)
\]

We know that for linear model $g^{lin}$:
\[
g^{lin}\left(\bm{w}(\xi K\tau+(r+1)\tau)\right)=\frac{1}{\sum_{i\in S(\xi,r)}n_i}\sum_{i\in S(\xi,r)}n_ig^{lin}\left(\bm{w}^{(i)}(\xi K\tau+r\tau+\tau)\right)
\]
However there will be a second order residual for a general nonlinear model:
\[
\begin{aligned}
&g\left(\bm{w}(\xi K\tau+(r+1)\tau)\right)\\
&=\frac{1}{\sum_{i\in S(\xi,r)}n_i}\sum_{i\in S(\xi,r)}n_ig\left(\bm{w}^{(i)}(\xi K\tau+r\tau+\tau)\right)-\eta_0^2\tau^2v_1^{(\xi,\tau)}+o\left(\tau^2\right)\\
\end{aligned}
\]
The residual term $v_1^{(\xi,\tau)}$ is defined as:
\[
v_1^{(\xi,\tau)}=\frac{g(\bm{w}(\xi K\tau+r\tau))^T}{\bar{n}l}\sum_{i\in\mathcal{S}(\xi,r)}n_i\left(D_1^{(\xi,r,i)}\right)^T\nabla^2gD_1^{(\xi,r,i)}g(\bm{w}(\xi K\tau+r\tau))
\]
It's norm is bounded by 
\[
\left\|v_1^{(\xi,\tau)}\right\|\le 8R_0^2\frac{C_1}{\sqrt{l}}C_2^2
\]
On the right hand side, we can use lemma \ref{lm::varofoperator} to derive:
\[
\begin{aligned}
&\frac{1}{\sum_{i\in S(\xi,r)}n_i}\sum_in_i\exp{\left(-\eta_0\frac{\Theta_{i0}}{n_i}t\right)}\left(g^{lin}(\bm{w}(\xi K\tau+r\tau))-g(\bm{w}(\xi K\tau+r\tau))\right)\\
&=\left(1+\eta_0^2\tau^2\text{Var}\left(\Theta_{i0}\right)\right)\text{exp}\left(-\eta_0\tau\frac{1}{\bar{n}}\sum_{i\in S(\xi,r)}\Theta_{i0}\right)\left(g^{lin}(\bm{w}(\xi K\tau+r\tau))-g(\bm{w}(\xi K\tau+r\tau))\right)\\
&+o\left(\eta_0^2\tau^2\right)
\end{aligned}
\]

As a result, the update can be written as:
\[
\begin{aligned}
&g^{lin}(\bm{w}(\xi K\tau+r\tau+\tau))-g(\bm{w}(\xi K\tau+r\tau+\tau))\\
&=\left(1+\eta_0^2\tau^2\text{Var}\left(\Theta_{i0}\right)\right)\text{exp}\left(-\eta_0\tau\frac{1}{\bar{n}}\sum_{i\in S(\xi,r)}\Theta_{i0}\right)\left(g^{lin}(\bm{w}(\xi K\tau+r\tau))-g(\bm{w}(\xi K\tau+r\tau))\right)\\
&+\eta^2\tau^2v_1^{(\xi,r)}\\
&+\frac{1}{\bar{n}}\sum_{i\in S(\xi,r)}n_i\exp{\left(-\eta_0\frac{\Theta_{i0}}{n_i}t\right)}\eta_0\int_0^{\tau}\exp{\left(\eta_0\frac{\Theta_{i0}}{n_i}s\right)}\left(\Theta_i(\bm{w}^{(i)}(\xi K\tau+r\tau+s))-\Theta_{i0}\right)g(s)ds\\
&+o(\tau^2)
\end{aligned}
\]
When we apply this relation iteratively for $r$, and still retain only leading order terms, we have:
\[
\begin{aligned}
&g^{lin}(\bm{w}_{\xi K\tau+K\tau+\tau})-g(\bm{w}_{\xi K\tau+K\tau+\tau})\\
&=\left(1+\eta_0^2\tau^2D_4^{(\xi,r)}\right)\text{exp}\left(-\eta_0\tau\frac{1}{\bar{n}}\Theta_{0}\right)\left(g^{lin}(\bm{w}(\xi K\tau))-g(\bm{w}(\xi K\tau))\right)\\
&+\sum_{i=1}^K\left(\prod_{j=K}^{i+1}\text{exp}\left(-\eta_0\tau\frac{\sum_{k\in S(\xi,j)}\Theta_{0j}}{\bar{n}}\right)\right)\Big(\eta_0^2\tau^2v_1^{(\xi,i)}\\
&+\frac{1}{\bar{n}}\sum_{k\in\mathcal{S}(\xi,i)}\exp{\left(-\eta_0\frac{\Theta_{k0}}{n_k}\tau\right)}\eta_0\int_0^{\tau}\exp{\left(\eta_0\frac{\Theta_{k0}}{n_k}s\right)}\left(\Theta_k(\bm{w}^{(k)}(\xi K\tau+r\tau+s))-\Theta_{k0})g(s)\right)ds\Big)\\
&+o(\tau^2)
\end{aligned}
\]
And we can take the norm on both sides:
\[
\begin{aligned}
&\left\|g^{lin}(\bm{w}(\xi K\tau+r\tau+\tau))-g(\bm{w}(\xi K\tau+r\tau+\tau))\right\|\\
&\le\left(1+\eta_0^2\tau^2\left\|D_4^{(\xi,r)}\right\|\right)\text{exp}\left(-\eta_0\tau\frac{1}{\bar{n}}\lambda_{min}\left(\Theta_{0}\right)\right)\left\|g^{lin}(\bm{w}(\xi K\tau+r\tau))-g(\bm{w}(\xi K\tau+r\tau))\right\|\\
&+\kappa_4+o(\tau^2)
\end{aligned}
\]
where $\kappa_4$ is defined as:
\[
\begin{aligned}
&\kappa_4=\exp \left(\frac{\eta_0\tau C_2^2}{\bar{n}}\right)\left[\eta_0^2\tau^28R_0^2C_1l^{-0.5}C_2^2+\eta_0\tau\exp \left(\frac{\eta_0\tau}{\bar{n}C_2^2}\right)2R_0C_2^2l^{-\frac{1}{2}}\right]\\
\end{aligned}
\]
From the expression above, we can see that $\kappa_4\to 0$ when $l\to\infty$. When $\eta_0\tau$ is small enough suth that $\left(1+\eta_0^2\tau^2\left\|D_4^{(\xi,r)}\right\|\right)\text{exp}\left(-\eta_0\tau\frac{1}{\bar{n}}\lambda_{min}\left(\Theta_{0}\right)\right)<1$, we have:
\[
\begin{aligned}
&\left\|g^{lin}(\bm{w}(\xi K\tau+r\tau+\tau))-g(\bm{w}(\xi K\tau+r\tau+\tau))\right\|\\
&\le \left(1-\left(1+\eta_0^2\tau^2\left\|D_4^{(\xi,r)}\right\|\right)\text{exp}\left(-\eta_0\tau\frac{1}{\bar{n}}\lambda_{min}\left(\Theta_{0}\right)\right)\right)^{-1}\kappa_4
\end{aligned}
\]

Thus the difference of error vector approaches zero in the limit $l\to\infty$, which indicates that $g^{lin}$ can approximate $g$ well.

\end{document}